\documentclass[10pt]{article} 
\usepackage[accepted]{tmlr}

\usepackage{amsmath,amsfonts,bm}

\def\eqref#1{equation~\ref{#1}}

\def\1{\bm{1}}

\DeclareMathAlphabet{\mathsfit}{\encodingdefault}{\sfdefault}{m}{sl}
\SetMathAlphabet{\mathsfit}{bold}{\encodingdefault}{\sfdefault}{bx}{n}

\usepackage{booktabs}       
\usepackage{amsfonts}       
\usepackage{nicefrac}       
\usepackage{microtype}      
\usepackage{xcolor}         
\usepackage{hyperref}
\usepackage{url}
\usepackage{graphicx}
\usepackage{amsmath}
\usepackage{amssymb}
\usepackage{mathtools}
\usepackage{amsthm}
\usepackage{algorithm}
\usepackage{algpseudocode}

\usepackage{multirow}
\usepackage[table]{xcolor}
\usepackage{caption}
\usepackage{float}
\usepackage{adjustbox}
\usepackage{subcaption}

\theoremstyle{plain}
\newtheorem{theorem}{Theorem}[section]

\theoremstyle{definition}

\theoremstyle{remark}

\title{CentroidKV: Efficient Long-Context LLM Inference
\\ via KV Cache Clustering}

\author{
    \textbf{Jie Hu}$^{1,2}$\thanks{Equal contribution.},
    \textbf{Shengnan Wang}$^{2}$\footnotemark[1],
    \textbf{Yutong He}$^{1}$,
    \textbf{Ping Gong}$^{3}$,
    \textbf{Jiawei Yi}$^{3}$,
    \textbf{Juncheng Zhang}$^{3}$, \\
    \textbf{Youhui Bai}$^{2}$,
    \textbf{Renhai Chen}$^{2}$,
    \textbf{Gong Zhang}$^{2}$,
    \textbf{Cheng Li}$^{3}$,
    \textbf{Kun Yuan}$^{1}$\thanks{Corresponding author.} \\
    {\normalfont\mdseries\itshape
    $^{1}$\textit{Peking University}
    \quad
    $^{2}$\textit{Huawei Technologies}
    \quad
    $^{3}$\textit{University of Science and Technology of China}}
}

\def\month{06}  
\def\year{2026} 
\def\openreview{\url{https://openreview.net/forum?id=T3EeupQhGj}} 

\begin{document}

\maketitle


\begin{abstract}
    Large language models (LLMs) with extended context windows have become increasingly prevalent for tackling complex tasks. However, the substantial Key-Value (KV) cache required for long-context LLMs poses significant deployment challenges. Existing approaches either discard potentially critical information needed for future generations or offer limited efficiency gains due to high computational overhead. In this paper, we introduce \textit{CentroidKV}, a simple yet effective framework for online KV cache clustering. Our approach is based on the observation that key states exhibit high similarity along the sequence dimension. To enable efficient clustering, we divide the sequence into chunks and propose \textit{Chunked Soft Matching}, which employs an alternating partition strategy within each chunk and identifies clusters based on similarity. CentroidKV then merges the KV cache within each cluster into a single centroid. Additionally, we provide a theoretical analysis of the computational complexity and the optimality of the intra-chunk partitioning strategy. Extensive experiments across various models and long-context benchmarks demonstrate that CentroidKV achieves up to 75\% reduction in KV cache memory usage while maintaining comparable model performance. Moreover, with minimal computational overhead, CentroidKV accelerates the decoding stage of inference by up to $1.92\times$ and increases the serving throughput by up to $4\times$.
\end{abstract}

\section{Introduction}

With the increasing demand to tackle a diverse range of complex real-world applications, such as multi-round dialogues, Large Language Models (LLMs) have been capable of supporting context windows of up to 1M tokens \citep{achiam2023gpt, touvron2023llama, team2024gemini}. However, deploying LLMs in long-context scenarios introduces substantial challenges, particularly related to the Key-Value (KV) cache. The KV cache stores the keys and values of all preceding tokens to avoid re-computation, and its memory requirements scale linearly with the context length. Due to the auto-regressive nature of LLMs, generating each token necessitates accessing the entire KV cache, making it a significant bottleneck for both inference latency and throughput. Moreover, the large size of KV cache imposes considerable demands on memory capacity, further complicating deployment.

To address these challenges, there is a pressing need for effective methods to reduce the KV cache size. Existing studies have explored this problem from multiple perspectives, including KV cache eviction \citep{zhang2023h2o, xiao2023efficient, ge2023model, li2024snapkv, liu2024scissorhands,yang2024pyramidinfer,cai2024pyramidkv}, KV cache merging \citep{zhangcam,  wang2024model, wan2024d2o}, quantization \citep{hooper2024kvquant, liu2024kivi}, and channel pruning \citep{xu2024think}, among others. By leveraging the inherent sparsity of the attention score matrix \citep{zhang2023h2o}, various methods have been proposed to reduce redundancy along the sequence length dimension. However, these approaches exhibit notable limitations. KV cache eviction, which discards less critical tokens based on historical attention scores, often leads to significant performance degradation. This occurs because tokens deemed unimportant in the current context may become crucial for future generations, especially in long-context scenarios. To improve the model performance after eviction, recent works \citep{yang2024pyramidinfer, cai2024pyramidkv, tang2024razorattention, feng2024ada, xiao2024duoattention, shi2024discovering} allocate different KV cache budget across layers and attention heads. Additionally, subsequent methods such as \citep{zhangcam} merge the tokens to be evicted with the tokens to be retained. However, existing methods typically rely on preliminary token eviction to define merging sets,  which often forces semantically distant tokens to be grouped together, resulting in centroid bias and information loss, ultimately degrading model performance. In contrast, our approach aims to compress the KV cache by clustering them into centroids based on the intrinsic similarity.

\begin{figure}[ht]
    \centering
    \begin{minipage}{0.485\linewidth}
        \centering
        \includegraphics[width=\linewidth]{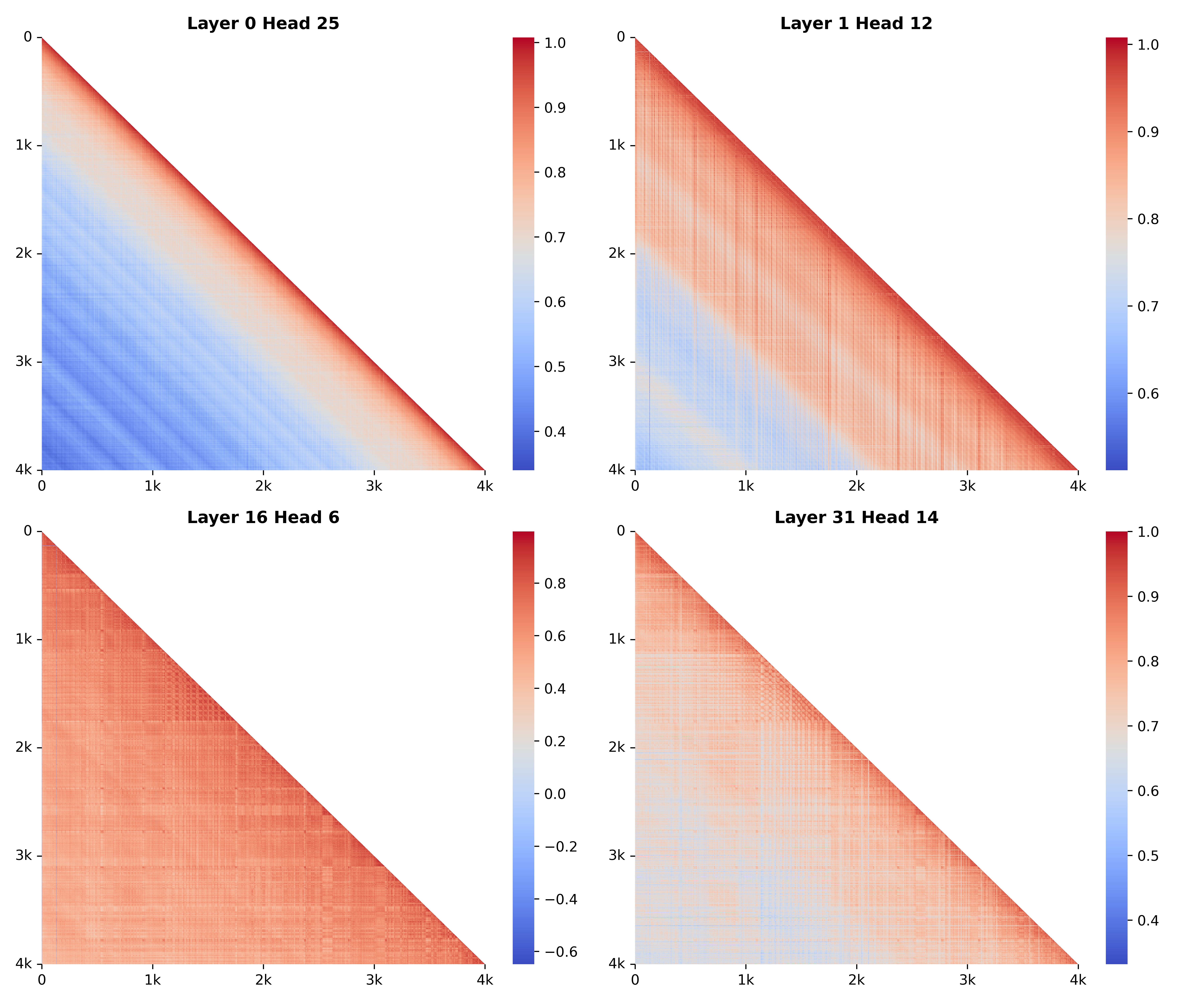}
        \caption{The cosine similarity maps of key states across various layers and heads.}
        \label{fig:obs_1}
    \end{minipage}
    \hfill
    \begin{minipage}{0.485\linewidth}
        \centering
        \includegraphics[width=\linewidth]{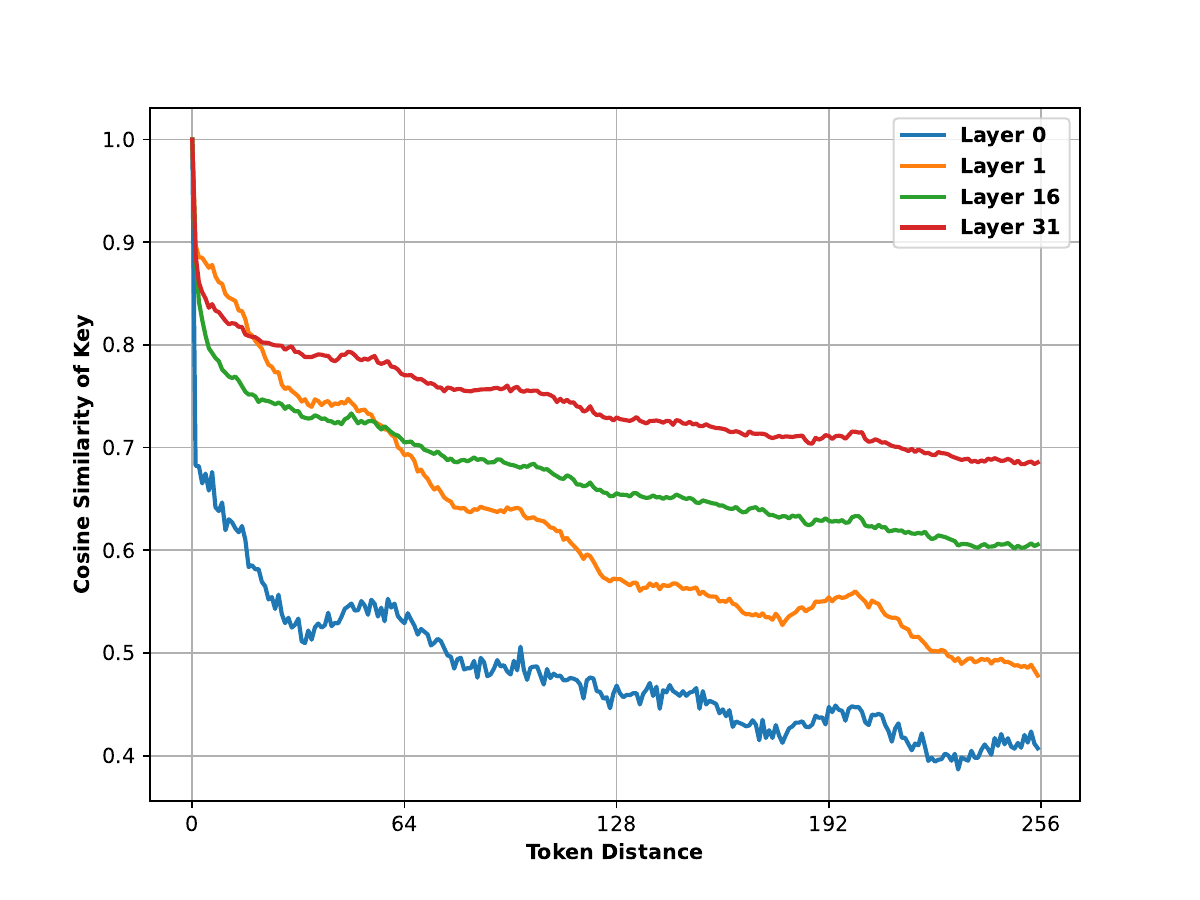}
        \caption{The correlation between token distance and the cosine similarity of key states.}
        \label{fig:obs_2}
    \end{minipage}
\end{figure}

Our approach is inspired by a key observation: key states exhibit high similarity along the sequence dimension, as illustrated in Figure~\ref{fig:obs_1}. This insight motivates us to cluster the KV cache by identifying merging sets solely based on token similarity. However, efficiently and accurately performing KV cache clustering online remains challenging, particularly given the long sequence lengths involved.

\begin{figure}[h]
\begin{center}
\includegraphics[width=0.9\columnwidth]{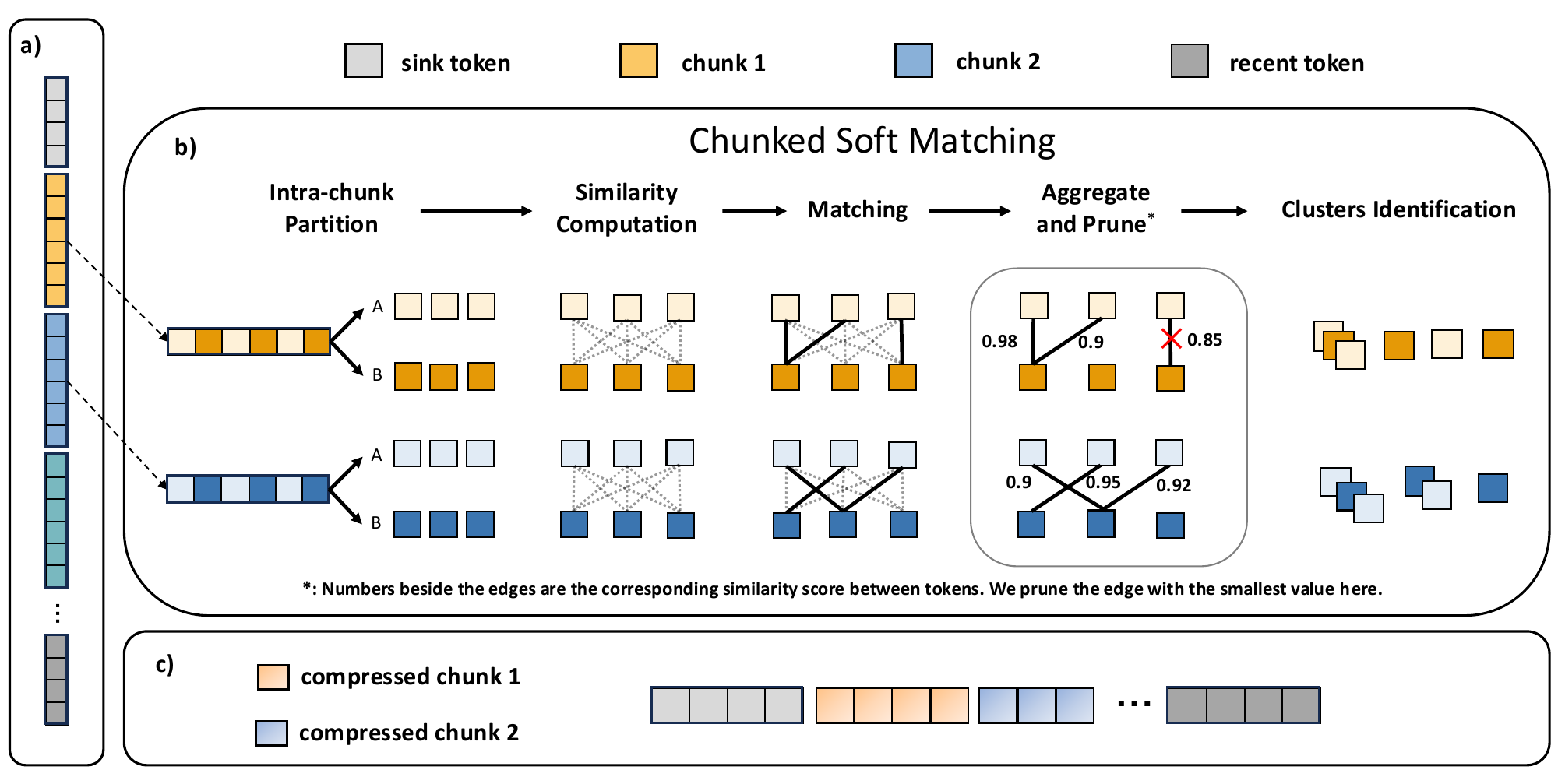}
\end{center}
\caption{An overview of CentroidKV: a) Divide sequences into chunks; b) Chunked Soft Matching to identify clusters; c) KV cache compression after clustering.}
\label{fig:CentroidKV}
\end{figure}

In this paper, we introduce \textit{CentroidKV}, a simple yet effective online KV cache clustering framework that improves the inference efficiency of LLMs in long-context scenarios. The core innovation of CentroidKV is the \textit{Chunked Soft Matching} algorithm, which enables efficient KV cache clustering with a favorable accuracy–efficiency trade-off. Inspired by the Bipartite Soft Matching algorithm \citep{bolya2022token} for Vision Transformers \citep{dosovitskiy2020image}, we are the first to extend this idea to the context of KV cache compression. Given the cache budget, one clustering round is executed as follows. The framework begins by dividing the sequence into chunks. Based on the observed correlation between token similarity and positional distance, we further partition each chunk in an alternating manner and theoretically prove the optimality of this intra-chunk partitioning strategy. Subsequently, \textit{Chunked Soft Matching} identifies clusters by locating highly similar token pairs across all chunks. Finally, the corresponding keys and values within each cluster are merged into a single centroid. As the decoding process advances, CentroidKV calls the clustering process if the cache size exceeds the budget.

We conduct extensive experiments on popular models to evaluate both the effectiveness and efficiency of CentroidKV. The results demonstrate that CentroidKV can reduce the KV cache memory usage by up to 75\% while maintaining comparable model performance. Furthermore, by dynamically clustering to preserve contextual information, CentroidKV outperforms baselines under a limited cache budget. Additionally, due to its simplicity and efficiency, CentroidKV accelerates the decoding stage of LLM inference by up to $1.92\times$ and increases the throughput by up to $4\times$ compared to full KV cache.

Our contributions are summarized as follows.
\begin{itemize}
    \item We introduce \textit{CentroidKV}, a simple yet effective framework for online KV cache clustering. The core innovation is a novel clustering algorithm, termed \textit{Chunked Soft Matching}.
    \item CentroidKV is a lightweight, plug-and-play solution to improve LLM inference efficiency. We theoretically analyze its computational complexity and formally prove the optimality of the intra-chunk partitioning strategy based on the observed similarity patterns.
    \item CentroidKV achieves up to a 75\% reduction in KV cache memory usage with minimal impact on model performance. Additionally, it accelerates the decoding stage by up to $1.92\times$ and increases the throughput by up to $4\times$.
\end{itemize}

\section{Related Work}


\paragraph{KV Cache Compression.}
Eviction-based approaches aim to maintain a fixed-size KV cache during decoding by selectively retaining informative tokens. H2O \citep{zhang2023h2o} retains a limited budget of KV cache by greedily discarding unimportant tokens based on the accumulated attention score. StreamingLLM \citep{xiao2023efficient} preserves the initial few tokens along with the recent tokens, based on the identification of attention sinks.  SnapKV \citep{li2024snapkv} selects important tokens for each attention head based on the observation window of prompts. FastGen \citep{ge2023model} conducts profiling for attention heads and dynamically evicts tokens based on different attention patterns. Despite their efficiency, these methods suffer from performance degradation in the long context scenario, since the information of the discarded token will be lost permanently. To mitigate this limitation, subsequent approaches \citep{zhangcam, wan2024d2o} introduce compensation mechanisms that merge evicted tokens into the remaining cache. However, their reliance on preliminary eviction still constrains their ability to preserve full contextual information. ClusterKV \citep{liu2024clusterkv} employs K-means clustering to enhance the retrieval of salient tokens; however, the high computational overhead of iterative clustering procedures limits its practicality for latency-sensitive online inference scenarios. Another line of work, such as PyramidKV \citep{cai2024pyramidkv} and PyramidInfer \citep{yang2024pyramidinfer}, focuses on optimizing cache budget allocation across layers to improve overall model accuracy.



\paragraph{Bipartite Soft Matching.} Introduced by ToMe \citep{bolya2022token}, Bipartite Soft Matching (BSM) is an algorithm designed to merge tokens in Vision Transformers (ViT) \citep{dosovitskiy2020image}, which is as fast as token pruning. Recent studies \citep{bolya2023token, kim2024token,tran2024accelerating} have extended BSM to enhance the efficiency of ViTs and Stable Diffusion \citep{rombach2022high} without compromising model performance.

\section{Preliminary}

In this section, we first give a basic preliminary about the KV cache clustering in LLM inference. For simplicity, we focus on a single attention head within a specific layer. 
LLM inference consists of two stages: pre-filling and decoding. During the pre-filling stage, the model generates the first token and initializes the KV cache, which stores the key and value states of the prompt as $K\in\mathbb{R}^{n\times d}$ and $V\in\mathbb{R}^{n\times d}$, respectively. In the decoding stage, for the given states of the current input $q,k,v\in\mathbb{R}^{1\times d}$, KV cache is updated as $K=\left[K, k\right]$, $V=\left[V, v\right]$. The vanilla attention output is then computed as follows:
\begin{align}
\vspace{-5pt}
\text{Attn}(q, K, V) = \text{softmax}\left(\frac{qK^T}{\sqrt{d_k}}\right)V 
=\frac{\sum_{i=1}^n \exp\left(\frac{q^T k_i}{\sqrt{d_k}}\right)v_i}{\sum_{i=1}^n \exp\left(\frac{q^T k_i}{\sqrt{d_k}}\right)}.
\vspace{-5pt}
\label{eq:attention}
\end{align}

For clearer derivation, we decompose KV cache into individual tokens, represented as $K=[k_1,\dots,k_n]$ and $V=[v_1,\dots,v_n]$, where each $k_i\in\mathbb{R}^{d}$ and $v_i\in\mathbb{R}^{d}$. In our approach, we utilize the cosine similarity as the distance metric between key states:
\begin{equation}
        \cos{\left(k_i,k_j\right)} = \frac{k_i\cdot k_j}{\Vert k_i \Vert\Vert k_j \Vert}
\end{equation}

Assume that tokens are grouped into clusters based on the similarity of their key states. The cluster centers are denoted as $\hat{K}=[\hat{k}_1,\dots,\hat{k}_C]$, where $C$ represents the number of clusters. The number of tokens in each cluster is denoted by $N=[n_1,\dots,n_C]$, with each cluster center $\hat{k}_t$ associated with a set of tokens $[\hat{k}_{1,t},\dots,\hat{k}_{n_t,t}]$, corresponding to the tokens in the $t$-th cluster. Here, $n_t$ is referred to as the cluster degree.

By approximating the attention output using the cluster centers in place of the original key states, the resulting attention output is as follows:
\begin{align*}
\vspace{-5pt}
\text{Attn}(q, K, V) &= \frac{\sum_{t=1}^C \sum_{i=1}^{n_t} \exp\left(\frac{q^T k_{i,t}}{\sqrt{d_k}}\right)v_{i,t}}{\sum_{t=1}^C \sum_{i=1}^{n_t} \exp\left(\frac{q^T k_{i,t}}{\sqrt{d_k}}\right)} \\
&\approx \frac{\sum_{t=1}^C \sum_{i=1}^{n_t} \exp\left(\frac{q^T \hat{k}_t}{\sqrt{d_k}}\right)v_{i,t}}{\sum_{t=1}^C \sum_{i=1}^{n_t} \exp\left(\frac{q^T \hat{k}_t}{\sqrt{d_k}}\right)} \\
&= \frac{\sum_{t=1}^C \left[\exp\left(\frac{q^T \hat{k}_t}{\sqrt{d_k}}\right) \sum_{i=1}^{n_t} v_{i,t}\right]}{\sum_{t=1}^C n_t \exp\left(\frac{q^T \hat{k}_t}{\sqrt{d_k}}\right)}.
\vspace{-5pt}
\end{align*}

By merging the value states corresponding to the same indices as the key states, namely $\hat{v}_t=\frac{\sum_{i=1}^{n_t} v_{i,t}}{n_t}$,  we formulate the approximate attention output, denoted as AppAttn, as follows:
\begin{align}
\vspace{-5pt}
\text{Attn}(q, K, V) 
&\approx \frac{\sum_{t=1}^C n_t \exp\left(\frac{q^T \hat{k}_t}{\sqrt{d_k}}\right)\hat{v}_t}{\sum_{t=1}^C n_t \exp\left(\frac{q^T \hat{k}_t}{\sqrt{d_k}}\right)} \nonumber\\
&= \frac{\sum_{t=1}^C \exp\left(\frac{q^T \hat{k}_t}{\sqrt{d_k}}+\log{n_t}\right)\hat{v}_t}{\sum_{t=1}^C \exp\left(\frac{q^T \hat{k}_t}{\sqrt{d_k}}+\log{n_t}\right)}
\vspace{-5pt} \nonumber\\
&\triangleq \text{AppAttn}(q, \hat{K}, \hat{V}, N),   
\label{eq:appattention}
\end{align}

where $\hat{V}=[\hat{v}_1,\dots,\hat{v}_C]$. Equation~(\ref{eq:appattention}) demonstrates that the approximate attention after clustering aligns with the vanilla attention in Equation~(\ref{eq:attention}) when each token is treated as a separate cluster. To minimize output error, it is essential that the key states within each cluster exhibit high similarity. Fortunately, the following observations further support the viability of this approach.

\section{Observations}
\label{sec:observation}

In this section, we present several empirical observations that motivate our approach.

\textbf{Observation 1. Key states exhibit high, localized similarity along the sequence dimension.}

Using the Llama-2-7B-32K model \citep{together2023llama2}, we randomly sample sequences of length 4K from the WikiText-2 \citep{merity2016pointer} dataset and perform zero-shot inference. The cosine similarity maps of the key states are visualized in Figure~\ref{fig:obs_1}. We observe that key states exhibit high cosine similarity between tokens across different layers and heads. Notably, tokens with high similarity tend to cluster within localized regions. These findings align with previous work \citep{wang2024model}.

This observation suggests that KV cache clustering can be leveraged for efficient inference without compromising accuracy. Furthermore, the observed localized similarity motivates us to subsequently enhance clustering efficiency by identifying similar tokens within local regions, rather than considering the entire sequence.


\textbf{Observation 2. As token distance increases, the cosine similarity of key states generally decreases monotonically and follows a convex trend.}

Building on the observed localized similarity, we further investigate the correlation between token distance and the cosine similarity of key states. We randomly sample multiple tokens within the sequence, define the local region as 256 tokens, and compute the average similarity across samples and attention heads. As illustrated in Figure~\ref{fig:obs_2}, we find that as token distance increases, the cosine similarity between key states generally follows a monotonically decreasing trend. Furthermore, this correlation appears to be convex with respect to the distance.

This observation motivates our framework design of a highly efficient clustering algorithm that minimizes the computational complexity of identifying similar token sets. Additionally, it provides empirical support for the theoretical analysis in Appendix~\ref{app:theory} which proves the optimality of the partitioning strategy in our framework.




\section{Method}

In this section, we introduce \textit{CentroidKV}, a simple yet effective framework for KV cache clustering designed to enhance the efficiency of long-context LLM inference. The framework comprises three key steps. First, the sequence is divided into chunks while preserving the attention sinks and recent tokens. Second, we propose a novel KV cache clustering algorithm, termed \textit{Chunked Soft Matching}, which employs an alternating partition strategy within each chunk and identifies clustering sets by finding highly similar token pairs across all chunks. Finally, the key and value states are merged based on the identified clusters, yielding a compressed KV cache for subsequent generations. An overview of CentroidKV is depicted in Figure~\ref{fig:CentroidKV}. 


\begin{algorithm}[tb]
\caption{Inference Pipeline with CentroidKV}
\label{alg:pipeline}
\begin{algorithmic}[1]
\Require cache ratio $R$, compression ratio $r$, maximum decoding length $\Gamma$, attention sink $n_1$, recent budget $n_2$, interval step $g$, chunk size $c$

\State \textbf{Pre-filling:} $Q,K,V\in\mathbb{R}^{n\times d}$
\State Cache length $s = n$, cluster degree $N = [1] \cdot n$, cache budget $B = R \cdot n$ \Comment{Initialization}
\State $O = \text{FlashAttn}(Q, K, V)$
\While{$s > B$}
    \State $K, V, N, s = \text{CentroidKV}(K, V, N, s, n_1, n_2, r, c)$ \Comment{First clustering after prefilling}
\EndWhile

\State \textbf{Decoding:} $q, k, v\in\mathbb{R}^{1\times d}$
\For{$i = 1, \dots, \Gamma-1$}
    \State $K=[K,k]$, $V=[V,v]$, $N=[N,1]$, $s = s + 1$ \Comment{Update the cache centroids}
    \State $O = \mathrm{softmax}(qK^T / \sqrt{d} + \log N)\cdot V$ \Comment{Equation~\ref{eq:appattention}}
    \If{$s \geq B + g$}
        \State $K, V, N, s = \text{CentroidKV}(K, V, N, s, n_1, n_2, r, c)$ \Comment{Clustering every $g$ decoding steps}
    \EndIf
\EndFor
\end{algorithmic}
\end{algorithm}

\subsection{Overall Inference Pipeline}

The overall pipeline of LLM inference integrated with CentroidKV is illustrated in Algorithm~\ref{alg:pipeline}. The cache budget $B$ is determined by the prompt length $n$ and the cache ratio $R$. In the pre-filling stage, after computing the attention output by Flash Attention \citep{dao2022flashattention}, CentroidKV is invoked to conduct the first clustering if the KV cache size exceeds the budget, which compresses the KV cache to centroids and records the cluster degree. In the decoding stage, the KV cache grows by one token at each step, causing the memory budget to be exceeded continuously. To amortize the compression overhead and maintain inference efficiency, CentroidKV is invoked periodically every $g$ decoding steps.

As illustrated in Figure~\ref{fig:CentroidKV}, CentroidKV primarily employs the Chunked Soft Matching algorithm to compress the KV cache. The compression ratio $r$ in Algorithm~\ref{alg:pipeline} governs the proportion of pruned edges in Figure~\ref{fig:CentroidKV}. Since Bipartite Soft Matching can reduce the cache size by at most half in a single pass, multiple rounds of clustering are performed to progressively compress the KV cache until it meets the predefined budget.

\subsection{Chunked Soft Matching}

Based on Observation 1 in Section~\ref{sec:observation}, CentroidKV begins by dividing the key states into chunks, as illustrated in Figure~\ref{fig:CentroidKV}. The localized similarity of key states ensures that this approach will not cause a significant loss of accuracy. We keep the attention sink and recent tokens before partitioning, regarding their importance for model performance \citep{zhang2023h2o, xiao2023efficient}.

Inspired by the Bipartite Soft Matching (BSM) algorithm introduced by \citep{bolya2022token} for token merging in the transformer block of Vision Transformers (ViT) \citep{dosovitskiy2020image}, we propose Chunked Soft Matching (CSM) for KV cache clustering in the second step of CentroidKV. BSM begins by partitioning the input tokens into two distinct sets, $A$ and $B$. Next, for each token in set $A$, an edge is drawn to its most similar token in set $B$. Among these edges, only the top-ranked similar connections are retained. Tokens that remain connected through these edges are then merged into a cluster, while others are left unchanged. Finally, the two sets, $A$ and $B$, are concatenated to form the output, incorporating both merged and unchanged tokens.

However, directly applying BSM for KV cache clustering poses two significant challenges. The first challenge arises from the large sequence dimension in long-context scenarios, which leads to inefficiencies in the matching process. The second challenge involves optimally partitioning the sequence into two sets, A and B, in a way that minimizes the impact on model accuracy.

\begin{table}
\caption{Computational complexity of distance matrix. $n$ is the sequence length and $d$ is hidden dimension. $k$ and $i$ of K-Means refer to number of cluster centers and iterations. $c$ of Chunked Soft Matching (CSM) refers to chunk size, which is relatively small to $n$.}
\label{tb:cmpofcomplexity}
\centering
\small
\begin{tabular}{cccc}
\toprule
Mesh & K-Means & BSM & CSM  \\
\midrule
$n^2d$ & $inkd$ & $\frac{1}{4}n^2d$ & $\frac{1}{4}ncd$ \\
\bottomrule
\end{tabular}
\end{table}

CSM addresses the first challenge through a preceding chunking step, which directly improves computational efficiency. While there are alternative approaches for KV cache clustering, such as K-means, they typically require multiple iterative updates to establish stable clusters, leading to substantial computational overhead. We summarize the computational complexity of representative methods in Table~\ref{tb:cmpofcomplexity}. In particular, the “Mesh” method computes pairwise cosine similarity across all tokens, incurring significant cost. In contrast, CSM achieves the lowest complexity by performing a single-pass clustering procedure with minimal token interactions.

Building on Observation 2 in Section~\ref{sec:observation}, we address the second challenge by partitioning each chunk into two sets, $A$ and $B$, in an alternating manner. The core idea of CSM is to assign highly similar states to different sets, ensuring that token pairs with high similarity are placed between sets, rather than within them. Furthermore, based on the observed monotonically decreasing and convex trend, we rigorously demonstrate the theoretical optimality of this intra-chunk partitioning strategy in Appendix~\ref{app:theory}. After computing the similarities, CSM aggregates the edges from all chunks and produces candidate matched pairs for KV cache merging.


\subsection{KV Cache Compression}


After candidate clusters are determined, CentroidKV performs selective merging to construct the compressed KV cache. Importantly, as shown in Figure~\ref{fig:CentroidKV}, not all matched pairs are merged. Instead, we rank all candidate matches by similarity and only merge the top fraction controlled by a compression ratio $r$. Concretely, $r<1.0$ retains only the top-$r$ most similar matches, discarding lower-confidence ones. This design avoids noisy aggregations from weakly similar tokens. As shown in Algorithm~\ref{alg:pipeline}, clustering is applied over multiple rounds to satisfy the cache budget. To further improve clustering quality across rounds, we adopt a progressively decreasing schedule for $r$, initialized at $r_{\text{init}}$ and linearly annealed with decay rate $\delta_r$:
\begin{equation}\label{eq:compression_ratio}
    r=r_{\text{init}}-j\cdot\delta_r
\end{equation}
where $j$ is the clustering round index. This schedule makes merging increasingly selective over time, preserving high-confidence structure in later rounds.

For each merged cluster, we maintain a degree counter $n_t$ to track its cumulative contribution across rounds. Both key and value states are aggregated using degree-weighted averaging. Specifically, for a cluster $k_1, k_2, \dots, k_{t}$ with corresponding degrees $n_1, n_2, \dots, n_{t}$, the centroid is computed as:
\begin{equation}
        \hat{k} = \frac{n_1k_1+n_2k_2+\dots+n_tk_{t}}{n_1+n_2+\dots+n_t}
\end{equation}
Value states are merged analogously using the same degree-based weighting. 

Finally, the resulting centroids are concatenated with the preserved attention sink and recent tokens to form the compressed KV cache for subsequent decoding steps.



\section{Experiments}
\label{sec:experiment}

In this section, we conduct comprehensive experiments to evaluate the effectiveness and efficiency of CentroidKV, followed by the ablation study on the framework design.

\subsection{Experimental Settings}

\paragraph{Models and baselines.} We employ two long-context models: Llama-3.1-8B-Instruct \citep{meta2024introducing} and Mistral-7B-Instruct-v0.2 \citep{jiang2023mistral7b}, serving as the backbone LLMs. We compare CentroidKV against state-of-the-art KV cache compression methods, including StreamingLLM \citep{xiao2023efficient}, SnapKV \citep{li2024snapkv} and PyramidKV \citep{cai2024pyramidkv}.

\paragraph{Datasets.} We evaluate CentroidKV using the widely recognized benchmarks: RULER \citep{hsieh2024ruler} and LongBench \citep{bai2023longbench}. As the variation of the Needle-in-a-Haystack test \citep{kamradt2024needle}, RULER includes 13 long-sequence tasks designed to assess the long-context understanding capabilities of LLMs. Additionally, LongBench includes tasks covering various application scenarios: single-document QA, multi-document QA, summarization, few-shot learning, synthetic tasks, and code completion.

\paragraph{Implementation Details.} All experiments are conducted on the KVPress codebase \citep{devoto2025expectedattention}, built upon HuggingFace Transformers \citep{wolf2019huggingface}. Following KVPress, each input is divided into context and question segments. In our evaluation protocol, only the context is available during compression, while the question is introduced afterward for answer generation. This setup more faithfully reflects real-world deployment, where future queries are unknown at compression time, and therefore constitutes a more challenging and realistic setting. 

Unless otherwise specified, CentroidKV is configured with 16 attention sinks and retains the 64 most recent tokens. The initial compression ratio is set to $r_{\text{init}}=0.8$ with a decay rate of $\delta_r=0.2$. The default chunk size is 256, and all computations are performed in bfloat16 precision. For experiments on RULER, the default context length is 4K tokens unless stated otherwise. Baseline methods are implemented following the configurations reported in their respective papers. Additional details are provided in Appendix~\ref{app:exp_set}. All experiments are conducted on high-performance GPUs delivering over 100 TFLOPS of compute.

\begin{figure*}[ht]
\centering
\begin{subfigure}[b]{0.48\textwidth}
\centering
\includegraphics[width=\textwidth]{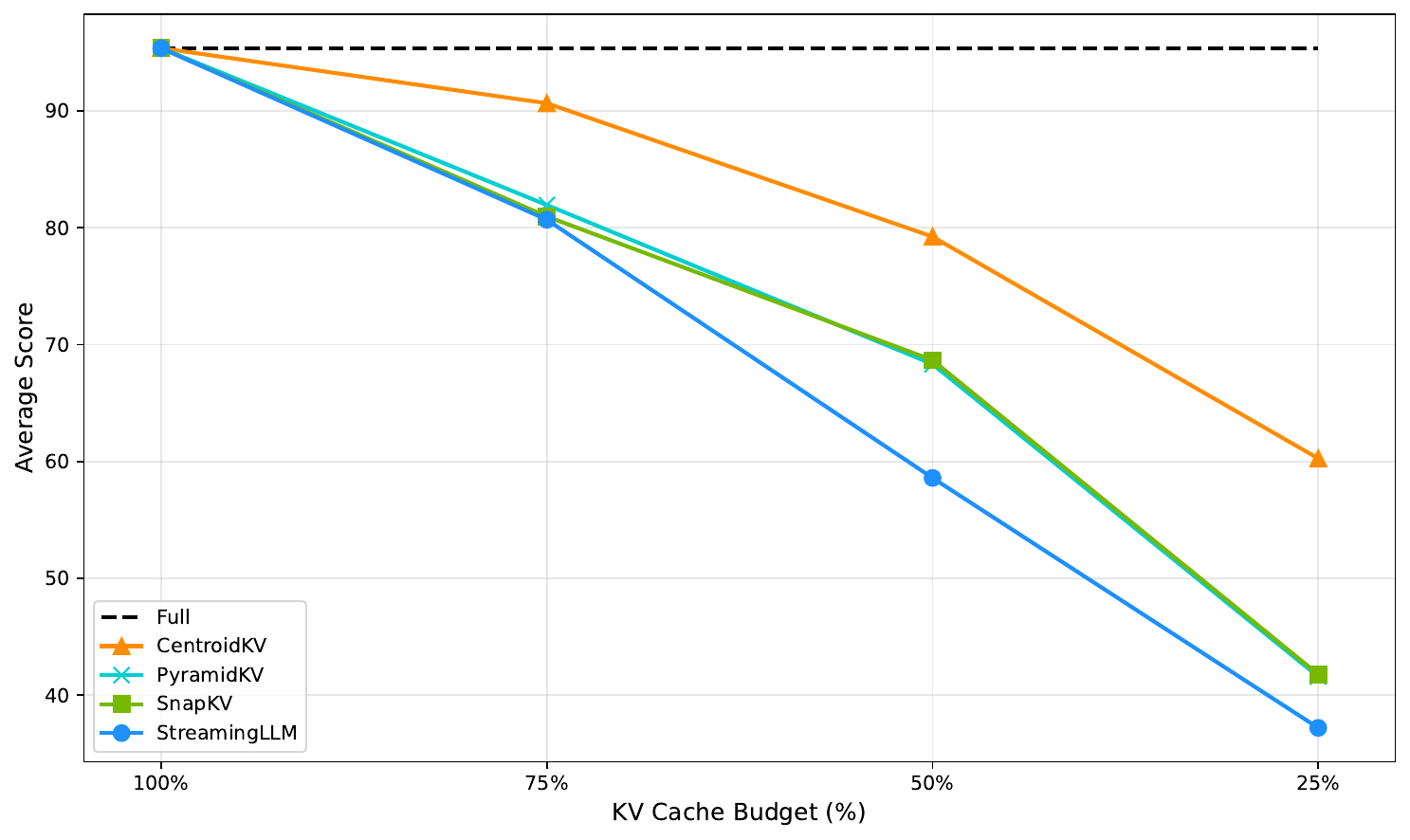}
\caption{Llama-3.1-8B-Instruct}
\label{fig:ruler_Llama_plot}
\end{subfigure}
\hfill
\begin{subfigure}[b]{0.48\textwidth}
\centering
\includegraphics[width=\textwidth]{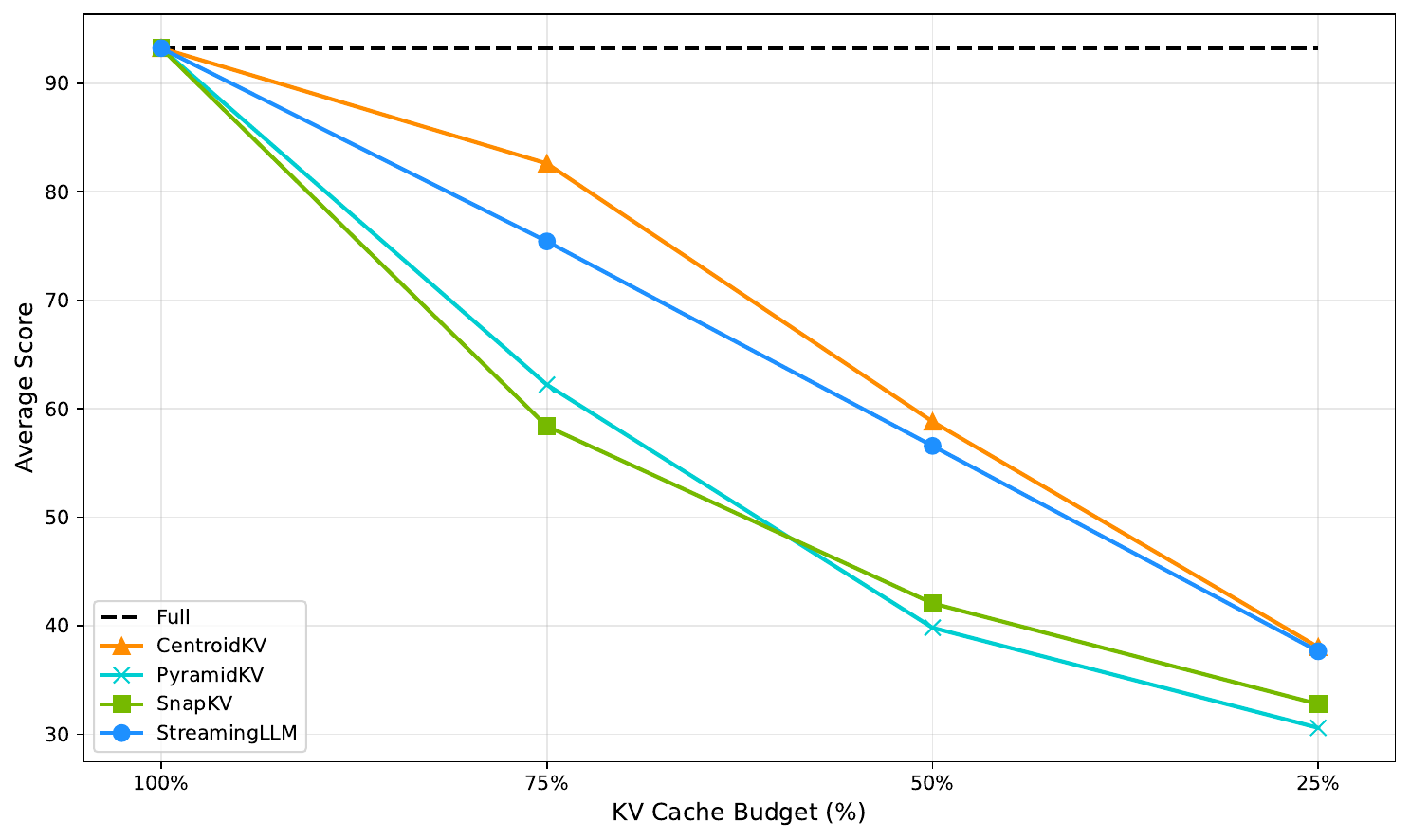}
\caption{Mistral-7B-Instruct-v0.2}
\label{fig:ruler_mistral_plot}
\end{subfigure}
\caption{Average score comparison of RULER benchmark across different cache budgets.}
\label{fig:ruler_plots}
\end{figure*}

\begin{table}[t]
\caption{Detailed comparison of RULER benchmark datasets across various budgets on Llama-3.1-8B-Instruct.}
\label{tab:ruler}
\begin{center}
\large
\begin{adjustbox}{max width=\columnwidth}
\begin{tabular}{clccccccccccccc}
\toprule
\textbf{Budget} & \textbf{Method} & \textbf{CWE} & \textbf{FWE} & \textbf{MK-NIAH-1} & \textbf{MK-NIAH-2} & \textbf{MK-NIAH-3} & \textbf{MQ-NIAH} & \textbf{MV-NIAH} & \textbf{S-NIAH-1} & \textbf{S-NIAH-2} & \textbf{S-NIAH-3} & \textbf{QA-1} & \textbf{QA-2} & \textbf{VT} \\
\midrule
100\% & Full & 99.31 & 95.80 & 100.00 & 100.00 & 100.00 & 100.00 & 100.00 & 100.00 & 100.00 & 100.00 & 88.24 & 56.67 & 99.77 \\
\midrule
\multirow{4}{*}{75\%} & StreamingLLM & \textbf{99.71} & 93.99 & 84.62 & 80.68 & \textbf{75.00} & 76.52 & 77.31 & 79.51 & 77.06 & 71.91 & \textbf{87.25} & 51.11 & 94.32 \\
 & SnapKV & 99.02 & 95.20 & 98.08 & 81.82 & 56.82 & 95.45 & 89.58 & 98.36 & 100.00 & 10.11 & 84.31 & 51.11 & 92.73 \\
 & PyramidKV & 99.51 & 93.69 & 98.08 & 85.23 & 53.41 & 98.23 & 97.92 & 99.18 & 100.00 & 10.11 & 86.27 & 47.78 & 95.91 \\
 & \cellcolor{cyan!10}CentroidKV & \cellcolor{cyan!10}99.12 & \cellcolor{cyan!10}\textbf{95.20} & \cellcolor{cyan!10}\textbf{100.00} & \cellcolor{cyan!10}\textbf{100.00} & \cellcolor{cyan!10}60.23 & \cellcolor{cyan!10}\textbf{99.75} & \cellcolor{cyan!10}\textbf{98.61} & \cellcolor{cyan!10}\textbf{100.00} & \cellcolor{cyan!10}\textbf{100.00} & \cellcolor{cyan!10}\textbf{87.64} & \cellcolor{cyan!10}84.31 & \cellcolor{cyan!10}\textbf{54.44} & \cellcolor{cyan!10}\textbf{99.32} \\
\midrule
\multirow{4}{*}{50\%} & StreamingLLM & 58.04 & 91.59 & 54.81 & 46.59 & \textbf{51.14} & 53.54 & 52.78 & 48.36 & 42.20 & \textbf{53.93} & \textbf{87.25} & 47.78 & 73.64 \\
 & SnapKV & \textbf{98.53} & 92.49 & 91.35 & 47.73 & 22.73 & 78.03 & 73.38 & 95.08 & 93.58 & 2.25 & 73.53 & 43.33 & 80.68 \\
 & PyramidKV & 88.63 & 90.39 & 85.58 & 57.95 & 18.18 & 76.01 & 77.78 & 96.72 & 97.25 & 1.12 & 73.53 & 43.33 & 81.82 \\
 & \cellcolor{cyan!10}CentroidKV & \cellcolor{cyan!10}97.45 & \cellcolor{cyan!10}\textbf{93.39} & \cellcolor{cyan!10}\textbf{99.04} & \cellcolor{cyan!10}\textbf{81.82} & \cellcolor{cyan!10}4.55 & \cellcolor{cyan!10}\textbf{97.47} & \cellcolor{cyan!10}\textbf{92.13} & \cellcolor{cyan!10}\textbf{100.00} & \cellcolor{cyan!10}\textbf{100.00} & \cellcolor{cyan!10}37.08 & \cellcolor{cyan!10}78.43 & \cellcolor{cyan!10}\textbf{52.22} & \cellcolor{cyan!10}\textbf{96.59} \\
\midrule
\multirow{4}{*}{25\%} & StreamingLLM & 10.29 & \textbf{93.39} & 32.69 & 22.73 & \textbf{21.59} & 29.04 & 26.85 & 27.87 & 19.27 & \textbf{28.09} & \textbf{89.22} & \textbf{38.89} & 43.64 \\
 & SnapKV & 86.08 & 85.59 & 33.65 & 10.23 & 1.14 & 24.75 & 24.07 & 80.33 & 53.21 & 1.12 & 55.88 & 27.78 & 59.32 \\
 & PyramidKV & \textbf{86.27} & 85.59 & 33.65 & 10.23 & 1.14 & 23.23 & 24.07 & 80.33 & 53.21 & 1.12 & 55.88 & 26.67 & 59.32 \\
 & \cellcolor{cyan!10}CentroidKV & \cellcolor{cyan!10}85.10 & \cellcolor{cyan!10}87.69 & \cellcolor{cyan!10}\textbf{87.50} & \cellcolor{cyan!10}\textbf{23.86} & \cellcolor{cyan!10}0.00 & \cellcolor{cyan!10}\textbf{77.27} & \cellcolor{cyan!10}\textbf{67.36} & \cellcolor{cyan!10}\textbf{100.00} & \cellcolor{cyan!10}\textbf{96.33} & \cellcolor{cyan!10}3.37 & \cellcolor{cyan!10}36.27 & \cellcolor{cyan!10}30.00 & \cellcolor{cyan!10}\textbf{88.64} \\
\bottomrule
\end{tabular}
\end{adjustbox}
\end{center}
\end{table}

\begin{table}[t]
\caption{Detailed comparison of RULER benchmark datasets across various budgets on Mistral-7B-Instruct.}
\label{tab:ruler_mistral}
\begin{center}
\large
\begin{adjustbox}{max width=\columnwidth}
\begin{tabular}{clccccccccccccc}
\toprule
\textbf{Budget} & \textbf{Method} & \textbf{CWE} & \textbf{FWE} & \textbf{MK-NIAH-1} & \textbf{MK-NIAH-2} & \textbf{MK-NIAH-3} & \textbf{MQ-NIAH} & \textbf{MV-NIAH} & \textbf{S-NIAH-1} & \textbf{S-NIAH-2} & \textbf{S-NIAH-3} & \textbf{QA-1} & \textbf{QA-2} & \textbf{VT} \\
\midrule
100\% & Full & 95.88 & 92.79 & 99.04 & 100.00 & 93.18 & 98.48 & 87.27 & 100.00 & 99.08 & 98.88 & 86.27 & 62.22 & 98.86 \\
\midrule
\multirow{4}{*}{75\%} & StreamingLLM & 92.75 & \textbf{91.89} & 84.62 & 80.68 & \textbf{68.18} & 65.91 & \textbf{73.38} & 79.51 & \textbf{77.06} & 70.79 & \textbf{86.27} & 56.67 & 52.73 \\
 & SnapKV & 95.00 & 90.99 & 38.46 & 76.14 & 67.05 & 32.32 & 25.00 & 87.70 & 67.89 & 6.74 & 85.29 & 58.89 & 27.50 \\
 & PyramidKV & \textbf{95.98} & 90.99 & 47.12 & 76.14 & 57.95 & 38.13 & 30.79 & 93.44 & 69.72 & 6.74 & 84.31 & \textbf{62.22} & 55.23 \\
 & \cellcolor{cyan!10}CentroidKV & \cellcolor{cyan!10}94.02 & \cellcolor{cyan!10}90.69 & \cellcolor{cyan!10}\textbf{87.50} & \cellcolor{cyan!10}\textbf{97.73} & \cellcolor{cyan!10}54.55 & \cellcolor{cyan!10}\textbf{79.80} & \cellcolor{cyan!10}65.97 & \cellcolor{cyan!10}\textbf{100.00} & \cellcolor{cyan!10}71.56 & \cellcolor{cyan!10}\textbf{96.63} & \cellcolor{cyan!10}83.33 & \cellcolor{cyan!10}53.33 & \cellcolor{cyan!10}\textbf{98.64} \\
\midrule
\multirow{4}{*}{50\%} & StreamingLLM & 90.98 & \textbf{89.79} & 54.81 & 46.59 & \textbf{40.91} & 33.08 & \textbf{52.08} & 46.72 & \textbf{42.20} & 52.81 & \textbf{87.25} & 47.78 & 50.45 \\
 & SnapKV & \textbf{94.31} & 88.89 & 23.08 & 44.32 & 17.05 & 15.40 & 14.58 & 78.69 & 22.02 & 2.25 & 76.47 & 51.11 & 18.64 \\
 & PyramidKV & 79.80 & 86.79 & 20.19 & 39.77 & 9.09 & 13.64 & 13.89 & 86.89 & 15.60 & 1.12 & 70.59 & \textbf{51.11} & 29.09 \\
 & \cellcolor{cyan!10}CentroidKV & \cellcolor{cyan!10}89.31 & \cellcolor{cyan!10}88.29 & \cellcolor{cyan!10}\textbf{57.69} & \cellcolor{cyan!10}\textbf{56.82} & \cellcolor{cyan!10}2.27 & \cellcolor{cyan!10}\textbf{34.09} & \cellcolor{cyan!10}29.86 & \cellcolor{cyan!10}\textbf{100.00} & \cellcolor{cyan!10}36.70 & \cellcolor{cyan!10}\textbf{53.93} & \cellcolor{cyan!10}71.57 & \cellcolor{cyan!10}46.67 & \cellcolor{cyan!10}\textbf{97.27} \\
\midrule
\multirow{4}{*}{25\%} & StreamingLLM & 60.20 & \textbf{90.99} & \textbf{32.69} & \textbf{21.59} & \textbf{18.18} & \textbf{16.67} & \textbf{26.16} & 27.87 & \textbf{19.27} & \textbf{28.09} & \textbf{88.24} & \textbf{40.00} & 19.32 \\
 & SnapKV & \textbf{87.35} & 84.08 & 14.42 & 14.77 & 0.00 & 13.89 & 14.35 & 69.67 & 11.01 & 1.12 & 60.78 & 38.89 & 15.68 \\
 & PyramidKV & 58.92 & 84.38 & 14.42 & 14.77 & 0.00 & 13.38 & 14.35 & 69.67 & 11.01 & 1.12 & 60.78 & 38.89 & 15.91 \\
 & \cellcolor{cyan!10}CentroidKV & \cellcolor{cyan!10}76.57 & \cellcolor{cyan!10}74.77 & \cellcolor{cyan!10}17.31 & \cellcolor{cyan!10}11.36 & \cellcolor{cyan!10}0.00 & \cellcolor{cyan!10}5.30 & \cellcolor{cyan!10}9.49 & \cellcolor{cyan!10}\textbf{100.00} & \cellcolor{cyan!10}15.60 & \cellcolor{cyan!10}3.37 & \cellcolor{cyan!10}51.96 & \cellcolor{cyan!10}34.44 & \cellcolor{cyan!10}\textbf{94.09} \\
\bottomrule
\end{tabular}
\end{adjustbox}
\end{center}
\end{table}

\begin{figure*}[ht]
\centering
\begin{subfigure}[b]{0.48\textwidth}
\centering
\includegraphics[width=\textwidth]{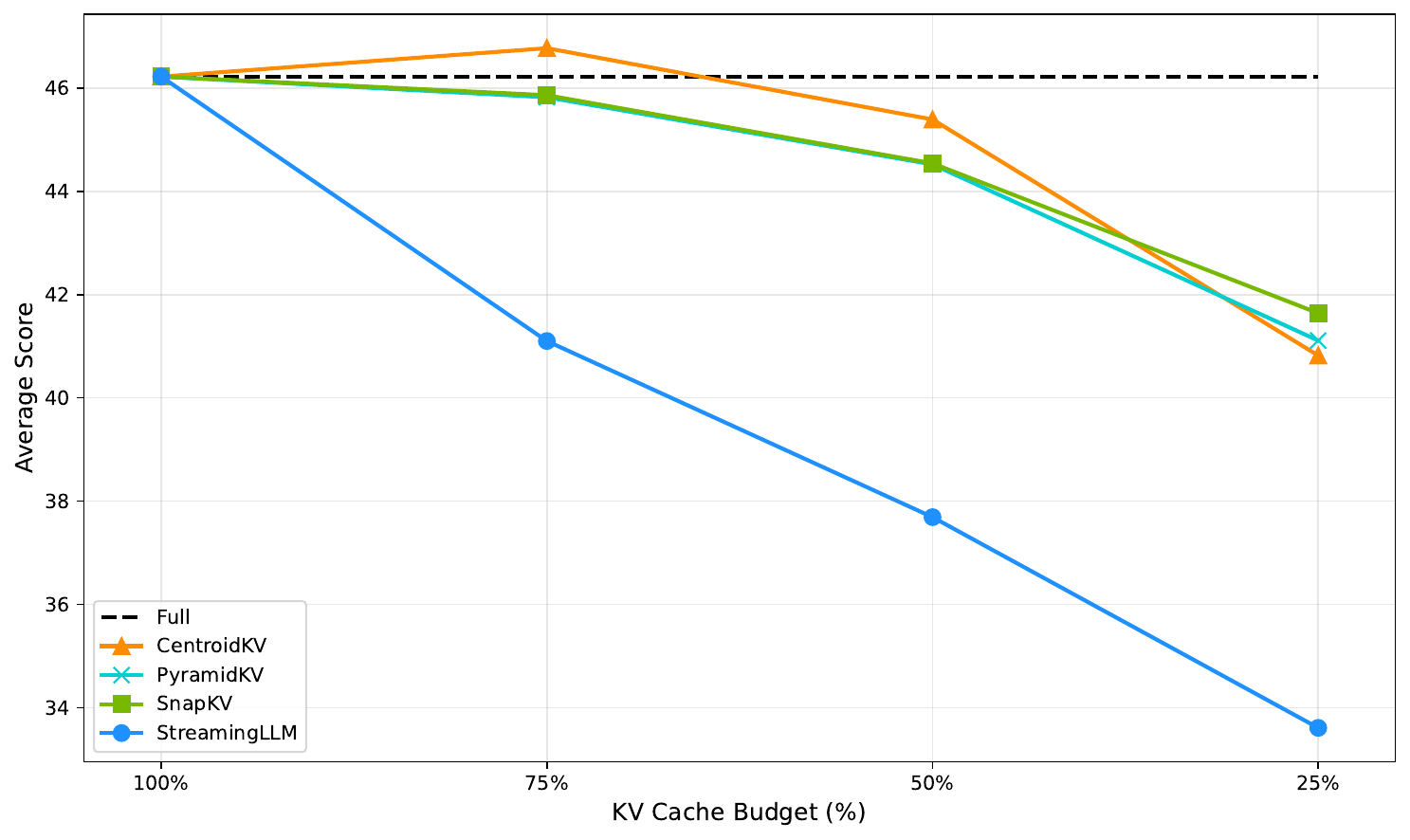}
\caption{Llama-3.1-8B-Instruct}
\label{fig:longbench_llama}
\end{subfigure}
\hfill
\begin{subfigure}[b]{0.48\textwidth}
\centering
\includegraphics[width=\textwidth]{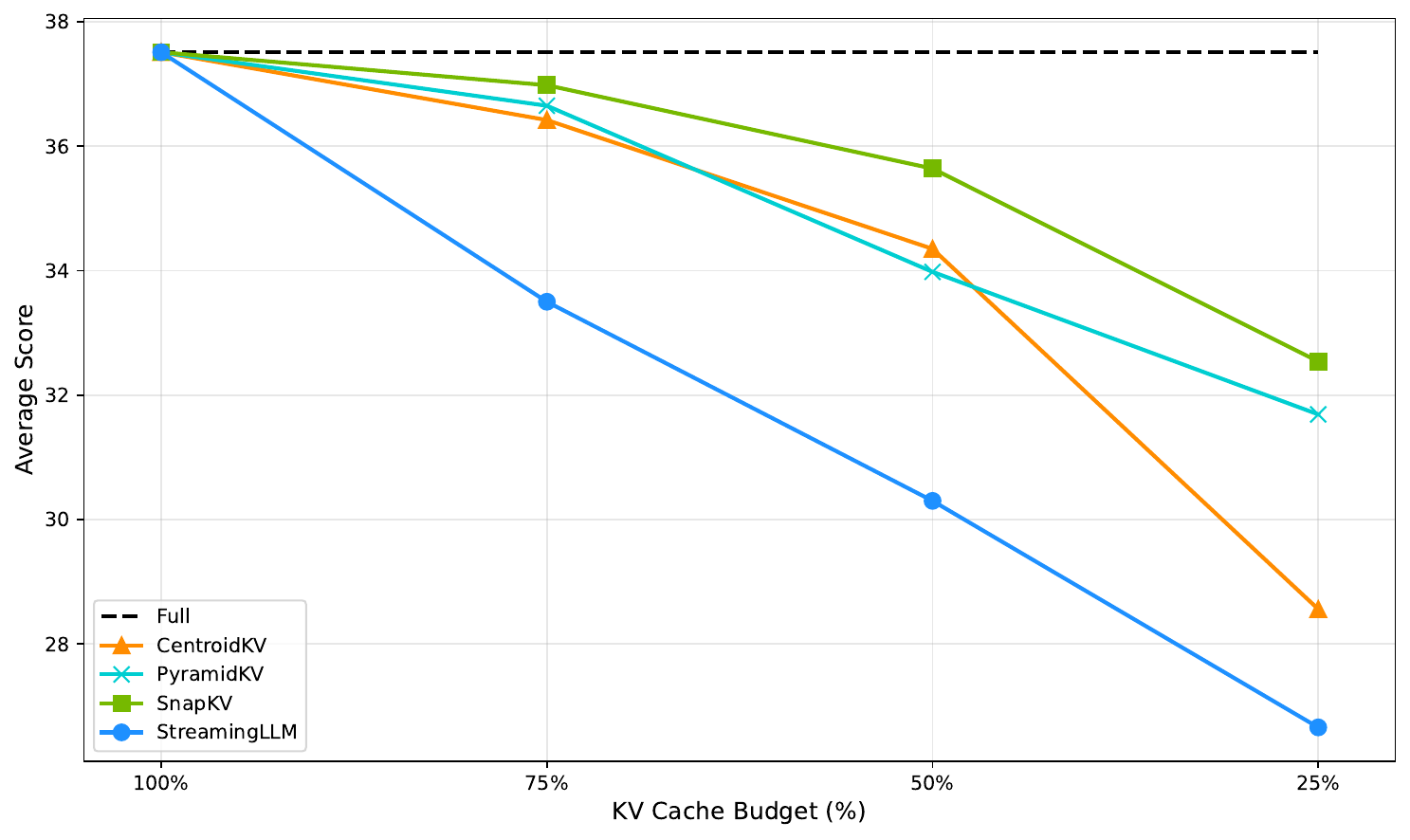}
\caption{Mistral-7B-Instruct-v0.2}
\label{fig:longbench_mistral}
\end{subfigure}
\caption{Average score comparison of LongBench datasets across different cache budgets.}
\label{fig:longbench_plots}
\end{figure*}


\begin{table}[t]
\caption{Task-average scores of LongBench across various budgets on Mistral-7B-Instruct.}
\label{tab:longbench_mistral_categories}
\begin{center}
\large
\begin{adjustbox}{max width=\columnwidth}
\begin{tabular}{clcccccc}
\toprule
\textbf{Budget} & \textbf{Method} & \textbf{Single-Doc QA} & \textbf{Multi-Doc QA} & \textbf{Summarization} & \textbf{Few-shot} & \textbf{Synthetic} & \textbf{Code} \\
\midrule
100\% & Full & 33.00 & 24.54 & 27.75 & 55.18 & 38.12 & 51.25 \\
\midrule
\multirow{4}{*}{75\%} & StreamingLLM & 25.92 & 22.96 & 26.80 & 51.73 & 28.20 & 48.72 \\
 & SnapKV & 31.23 & 23.92 & 27.23 & 54.55 & 39.01 & 51.41 \\
 & PyramidKV & 30.71 & 23.92 & 27.19 & 53.36 & \textbf{39.32} & 51.10 \\
 & \cellcolor{cyan!10}CentroidKV & \cellcolor{cyan!10}\textbf{31.34} & \cellcolor{cyan!10}\textbf{24.04} & \cellcolor{cyan!10}\textbf{27.54} & \cellcolor{cyan!10}\textbf{54.95} & \cellcolor{cyan!10}33.15 & \cellcolor{cyan!10}\textbf{51.45} \\
\midrule
\multirow{4}{*}{50\%} & StreamingLLM & 22.92 & 19.52 & 25.90 & 48.87 & 18.27 & 48.30 \\
 & SnapKV & 26.91 & \textbf{22.24} & 25.96 & \textbf{54.34} & \textbf{39.23} & \textbf{51.70} \\
 & PyramidKV & 25.70 & 18.17 & 25.23 & 52.55 & 38.70 & 50.73 \\
 & \cellcolor{cyan!10}CentroidKV & \cellcolor{cyan!10}\textbf{29.63} & \cellcolor{cyan!10}20.91 & \cellcolor{cyan!10}\textbf{26.37} & \cellcolor{cyan!10}51.94 & \cellcolor{cyan!10}31.05 & \cellcolor{cyan!10}50.49 \\
\midrule
\multirow{4}{*}{25\%} 
 & StreamingLLM & 19.05 & 15.95 & 23.94 & 44.10 & 10.79 & 47.96 \\
 & SnapKV & 21.54 & 16.34 & 24.01 & \textbf{53.29} & \textbf{36.36} & \textbf{51.16} \\
 & PyramidKV & 20.71 & 15.53 & 23.61 & 51.94 & 35.08 & 50.77 \\
 & \cellcolor{cyan!10}CentroidKV & \cellcolor{cyan!10}\textbf{23.02} & \cellcolor{cyan!10}\textbf{17.09} & \cellcolor{cyan!10}\textbf{24.47} & \cellcolor{cyan!10}42.50 & \cellcolor{cyan!10}19.13 & \cellcolor{cyan!10}48.72 \\
\bottomrule
\end{tabular}
\end{adjustbox}
\end{center}
\end{table}

\begin{table}[t]
\caption{Task-average scores of LongBench across various budgets on Llama-3.1-8B-Instruct.}
\label{tab:longbench_llama_categories}
\begin{center}
\large
\begin{adjustbox}{max width=\columnwidth}
\begin{tabular}{clcccccc}
\toprule
\textbf{Budget} & \textbf{Method} & \textbf{Single-Doc QA} & \textbf{Multi-Doc QA} & \textbf{Summarization} & \textbf{Few-shot} & \textbf{Synthetic} & \textbf{Code} \\
\midrule
100\% & Full & 44.90 & 48.21 & 29.17 & 53.83 & 55.35 & 50.27 \\
\midrule
\multirow{4}{*}{75\%} & StreamingLLM & 36.18 & 42.01 & 27.53 & 53.87 & 42.27 & 47.13 \\
 & SnapKV & 43.78 & 47.38 & 28.36 & 54.28 & 55.35 & \textbf{50.86} \\
 & PyramidKV & 44.02 & 47.32 & 28.25 & 54.22 & 55.35 & 50.48 \\
 & \cellcolor{cyan!10}CentroidKV & \cellcolor{cyan!10}\textbf{44.86} & \cellcolor{cyan!10}\textbf{47.40} & \cellcolor{cyan!10}\textbf{28.49} & \cellcolor{cyan!10}\textbf{57.11} & \cellcolor{cyan!10}\textbf{56.58} & \cellcolor{cyan!10}50.80 \\
\midrule
\multirow{4}{*}{50\%} & StreamingLLM & 31.27 & 37.54 & 26.08 & 53.62 & 29.98 & 48.80 \\
 & SnapKV & 38.73 & 45.20 & 26.96 & 56.27 & 55.27 & 50.28 \\
 & PyramidKV & 39.63 & \textbf{45.38} & \textbf{27.18} & 54.52 & 54.75 & \textbf{51.31} \\
 & \cellcolor{cyan!10}CentroidKV & \cellcolor{cyan!10}\textbf{42.71} & \cellcolor{cyan!10}43.66 & \cellcolor{cyan!10}27.04 & \cellcolor{cyan!10}\textbf{57.57} & \cellcolor{cyan!10}\textbf{55.75} & \cellcolor{cyan!10}50.86 \\
\midrule
\multirow{4}{*}{25\%} 
  & StreamingLLM & 24.42 & 29.98 & 24.46 & 53.11 & 20.25 & 50.67 \\
 & SnapKV & 31.60 & \textbf{41.93} & \textbf{25.00} & \textbf{56.66} & 49.55 & \textbf{50.77} \\
 & PyramidKV & 30.60 & 40.05 & 24.61 & 56.00 & \textbf{51.52} & 50.45 \\
 & \cellcolor{cyan!10}CentroidKV & \cellcolor{cyan!10}\textbf{33.37} & \cellcolor{cyan!10}36.95 & \cellcolor{cyan!10}24.59 & \cellcolor{cyan!10}55.29 & \cellcolor{cyan!10}51.00 & \cellcolor{cyan!10}50.25 \\
\bottomrule
\end{tabular}
\end{adjustbox}
\end{center}
\end{table}

\subsection{Accuracy Evaluation}

We evaluate accuracy under varying KV cache budgets on RULER and LongBench, reporting results at 25\%--75\% of the full cache to characterize the accuracy–compression trade-off.

\paragraph{RULER.}
Figure~\ref{fig:ruler_plots} shows average performance over 13 RULER datasets. CentroidKV achieves the strongest overall performance on both Llama-3.1-8B-Instruct and Mistral-7B-Instruct-v0.2. Notably, performance gaps widen as the cache budget decreases, highlighting the robustness of CentroidKV under aggressive compression. Detailed results in Tables~\ref{tab:ruler} and~\ref{tab:ruler_mistral} corroborate these observations: CentroidKV attains the best or near-best performance on most datasets and budgets, particularly on S-NIAH-1/2, MK-NIAH-1, MQ-NIAH, and VT. Performance degradation is observed on S-NIAH-3 and MK-NIAH-3, which involve UUID-type keys and values consisting of semantically arbitrary random strings. Because CentroidKV clusters tokens based on cosine similarity of key states, it is effective at preserving tokens with distinctive semantic representations (e.g., words or numbers), but struggles to differentiate tokens that lack meaningful semantic structure. As a result, UUID tokens are not reliably retained. This limitation is shared by other content-aware methods such as SnapKV and PyramidKV, whereas StreamingLLM remains stable by preserving tokens based on fixed positional rules rather than content.

\paragraph{LongBench.}
Figure~\ref{fig:longbench_plots} reports average results over 16 LongBench datasets. To provide a more fine-grained analysis, we further group datasets into six task categories (Appendix~\ref{app:longbench_detail}) and report macro-averaged results in Tables~\ref{tab:longbench_mistral_categories} and~\ref{tab:longbench_llama_categories}. CentroidKV performs consistently well on Single-Doc QA, Multi-Doc QA, and Summarization, achieving the best or near-best category-level averages under most budgets for both models. In contrast, performance drops are observed on Synthetic tasks and certain few-shot retrieval-style tasks, particularly on Mistral. Per-dataset results at the 25\% budget (Tables~\ref{tab:longbench_llama_25pct} and~\ref{tab:longbench_mistral_25pct}) show that these degradations are concentrated in fine-grained retrieval and index-sensitive tasks. In such settings, aggressive token merging can obscure exact token matches and positional signals that are critical for retrieval-based queries. By contrast, CentroidKV remains highly effective on tasks requiring semantic aggregation and reasoning, where preserving high-level contextual information is more important than exact token fidelity.

\paragraph{Summary.}
Overall, CentroidKV delivers strong accuracy across both benchmarks, particularly under constrained KV cache budgets. Its strengths are most evident in tasks requiring semantic understanding, while its limitations arise in retrieval-intensive scenarios that depend on precise token-level information.

\begin{table}
\caption{Latency and memory comparison across varying context lengths on Llama-3.1-8B-Instruct.}
\footnotesize
\label{tab:eff_lm3_all}
\centering
\begin{small}
\begin{adjustbox}{max width=\textwidth}
\begin{tabular}{clcccc}
   \toprule
    Context Length & Method & TTFT (s) & TPOT (ms) & Speedup $\uparrow$ & KV Cache (GB)\\
    \midrule
    \multirow{5}{*}{32k} & Full & 4.006 & 43.64 & 1$\times$ & 4.03 \\
    \cmidrule(lr){2-6}
    & SnapKV & 4.096 & 35.21 & 1.24$\times$ & 1.10 \\
    & StreamingLLM & 4.000 & 35.08 & 1.24$\times$ & 1.10 \\
    & PyramidKV & 4.109 & 35.48 & 1.23$\times$ & 1.10 \\
    & \cellcolor{cyan!10}CentroidKV & \cellcolor{cyan!10}4.267 & \cellcolor{cyan!10}\textbf{34.72} & \cellcolor{cyan!10}\textbf{1.26$\times$} & \cellcolor{cyan!10}1.10 \\
    \midrule
    \multirow{5}{*}{64k} & Full & 10.837 & 57.29 & 1$\times$ & 7.93 \\
    \cmidrule(lr){2-6}
    & SnapKV & 11.057 & 37.10 & 1.54$\times$ & 2.08 \\
    & StreamingLLM & 10.860 & \textbf{36.87} & \textbf{1.55$\times$} & 2.08 \\
    & PyramidKV & 11.057 & 38.11 & 1.50$\times$ & 2.08 \\
    & \cellcolor{cyan!10}CentroidKV & \cellcolor{cyan!10}11.291 & \cellcolor{cyan!10}37.87 & \cellcolor{cyan!10}1.51$\times$ & \cellcolor{cyan!10}2.08 \\
    \midrule
    \multirow{5}{*}{128k} & Full & 34.938 & 85.12 & 1$\times$ & 15.75 \\
    \cmidrule(lr){2-6}
    & SnapKV & 34.551 & 44.32 & 1.92$\times$ & 4.03 \\
    & StreamingLLM & 34.145 & \textbf{44.30} & \textbf{1.92$\times$} & 4.03 \\
    & PyramidKV & 34.532 & 44.42 & 1.92$\times$ & 4.03 \\
    & \cellcolor{cyan!10}CentroidKV & \cellcolor{cyan!10}35.001 & \cellcolor{cyan!10}44.57 & \cellcolor{cyan!10}1.91$\times$ & \cellcolor{cyan!10}4.03 \\
   \bottomrule
\end{tabular}
\end{adjustbox}
\end{small}
\end{table}

\begin{table}
\caption{Latency and memory comparison across varying context lengths on Mistral-7B-Instruct.}
\footnotesize
\label{tab:eff_mistral_all}
\centering
\begin{small}
\begin{adjustbox}{max width=\textwidth}
\begin{tabular}{clcccc}
   \toprule
    Context Length & Method & TTFT (s) & TPOT (ms) & Speedup $\uparrow$ & KV Cache (GB)\\
    \midrule
    \multirow{5}{*}{32k} & Full & 3.920 & 43.42 & 1$\times$ & 4.03 \\
    \cmidrule(lr){2-6}
    & SnapKV & 4.027 & \textbf{34.23} & \textbf{1.27$\times$} & 1.10 \\
    & StreamingLLM & 3.918 & 34.95 & 1.24$\times$ & 1.10 \\
    & PyramidKV & 3.961 & 35.31 & 1.23$\times$ & 1.10 \\
    & \cellcolor{cyan!10}CentroidKV & \cellcolor{cyan!10}4.184 & \cellcolor{cyan!10}34.79 & \cellcolor{cyan!10}1.25$\times$ & \cellcolor{cyan!10}1.10 \\
    \midrule
    \multirow{5}{*}{64k} & Full & 10.650 & 57.09 & 1$\times$ & 7.93 \\
    \cmidrule(lr){2-6}
    & SnapKV & 10.872 & 37.12 & 1.54$\times$ & 2.08 \\
    & StreamingLLM & 10.637 & \textbf{36.98} & \textbf{1.54$\times$} & 2.08 \\
    & PyramidKV & 10.884 & 37.49 & 1.52$\times$ & 2.08 \\
    & \cellcolor{cyan!10}CentroidKV & \cellcolor{cyan!10}11.111 & \cellcolor{cyan!10}37.23 & \cellcolor{cyan!10}1.53$\times$ & \cellcolor{cyan!10}2.08 \\
    \midrule
    \multirow{5}{*}{128k} & Full & 34.661 & 84.71 & 1$\times$ & 15.75 \\
    \cmidrule(lr){2-6}
    & SnapKV & 33.897 & 44.19 & 1.92$\times$ & 4.03 \\
    & StreamingLLM & 33.802 & 44.54 & 1.90$\times$ & 4.03 \\
    & PyramidKV & 34.182 & 44.50 & 1.90$\times$ & 4.03 \\
    & \cellcolor{cyan!10}CentroidKV & \cellcolor{cyan!10}34.668 & \cellcolor{cyan!10}\textbf{44.12} & \cellcolor{cyan!10}\textbf{1.92$\times$} & \cellcolor{cyan!10}4.03 \\
   \bottomrule
\end{tabular}
\end{adjustbox}
\end{small}
\end{table}

\subsection{Efficiency Evaluation}

We evaluate inference efficiency in terms of decoding latency and GPU memory footprint. Following standard practice, we report Time To First Token (TTFT) to capture pre-filling overhead, and Time Per Output Token (TPOT) to measure decoding efficiency. All compression-based methods are configured with a uniform 25\% KV cache budget and evaluated against the full-attention baseline under an identical inference pipeline. In addition, we integrate CentroidKV into the vLLM serving framework and report the results in Appendix~\ref{app:vllm}, where CentroidKV improves serving throughput by up to $4.0\times$.

\paragraph{Latency.}
Tables~\ref{tab:eff_lm3_all} and~\ref{tab:eff_mistral_all} present latency across context lengths ranging from 32K to 128K, with a fixed decoding length of 1K tokens. Overall, CentroidKV achieves competitive decoding efficiency while incurring only modest pre-filling overhead. Specifically, TTFT is slightly higher than full baseline and prior compression methods due to the additional online clustering operations. Nevertheless, the overhead remains consistently small across all evaluated settings. During decoding, CentroidKV achieves speedups comparable to prior compression methods, achieving up to $1.91\times$ TPOT reduction on Llama and $1.92\times$ on Mistral at 128K context length. This is as expected since all methods operate under the same effective KV cache budget.

\paragraph{Memory.}
As shown in Tables~\ref{tab:eff_lm3_all} and~\ref{tab:eff_mistral_all}, all compression-based methods, including CentroidKV, reduce KV cache memory by approximately $4\times$ compared to full attention. These results demonstrate that CentroidKV achieves memory efficiency comparable to prior methods.

\paragraph{Summary.}
Overall, CentroidKV achieves competitive latency and memory efficiency compared to prior compression-based methods, while consistently delivering improved accuracy under the same KV cache budget.



\begin{table}
\caption{Ablation study (RULER score and TTFT(s)) on chunking strategy and varied chunk size.}
\footnotesize
\label{tab:ablation_chunk}
\centering
\begin{small}
\begin{tabular}{lcccccc}
\toprule
Chunk Size & Avg. Score & 16k  & 32k & 64k & 128k \\
\midrule
64 & 59.76 & 1.791 & 4.195 & 11.316 & 34.849 \\
128 & 59.06 & 1.792 & 4.268 & 11.284 & 34.856 \\
\cellcolor{cyan!10}256 & \cellcolor{cyan!10}59.87 & \cellcolor{cyan!10}1.815 & \cellcolor{cyan!10}4.257 & \cellcolor{cyan!10}11.301 & \cellcolor{cyan!10}34.896 \\
w/o chunk & 60.62 & 1.951 & 4.941 & 14.415 & OOM \\
\bottomrule
\end{tabular}
\end{small}
\end{table}

\begin{table}[t]
\centering
\caption{Ablation study (RULER score and TTFT(s)) on scheduling of compression ratio defined in Eq.~\ref{eq:compression_ratio}.}
\label{tab:ablation_rate}
\begin{small}
\begin{tabular}{ccccccc}
\toprule
$r_{\text{init}}$  & $\delta_r$ & Avg. Score & 16k & 32k & 64k & 128k \\
\midrule
\multirow{4}{*}{1.00} & 0.00 & 19.22 & 1.715 & 4.110 & 11.105 & 34.601 \\
 & 0.10 & 23.77 & 1.720 & 4.113 & 11.106 & 34.650 \\
 & 0.20 & 25.80 & 1.718 & 4.122 & 11.092 & 34.674 \\
 & 0.30 & 26.69 & 1.754 & 4.142 & 11.129 & 34.717 \\
\midrule
\multirow{4}{*}{0.90} & 0.00 & 47.29 & 1.717 & 4.114 & 11.077 & 34.648 \\
 & 0.10 & 56.25 & 1.720 & 4.118 & 11.098 & 34.606 \\
 & 0.20 & 59.26 & 1.734 & 4.139 & 11.157 & 34.779 \\
 & 0.30 & 56.77 & 1.761 & 4.170 & 11.223 & 34.777 \\
\midrule
\multirow{4}{*}{0.80} & 0.00 & 51.44 & 1.723 & 4.186 & 11.088 & 34.688 \\
 & 0.10 & 58.97 & 1.738 & 4.116 & 11.152 & 34.710 \\
 & \cellcolor{cyan!10}0.20 & \cellcolor{cyan!10}60.46 & \cellcolor{cyan!10}1.786 & \cellcolor{cyan!10}4.168 & \cellcolor{cyan!10}11.274 & \cellcolor{cyan!10}34.910 \\
 & 0.30 & 57.33 & 1.840 & 4.264 & 11.364 & 35.100 \\
\bottomrule
\end{tabular}
\end{small}
\end{table}






\subsection{Ablation Study}




We conduct ablations on both accuracy and efficiency to analyze the key design choices in CentroidKV, focusing on chunked clustering strategy and the scheduling of compression ratio. Results are reported using average accuracy on RULER and TTFT across varying context lengths.

\paragraph{Chunking.}
Table~\ref{tab:ablation_chunk} evaluates the impact of chunk size $c$ in Algorithm~\ref{alg:pipeline}. Without chunking, clustering is performed over the full sequence, leading to the highest accuracy but incurs substantial overhead and fails to scale to long contexts (OOM at 128K). Introducing chunking reduces the matching scope from the entire sequence to local segments, significantly lowering latency and improving scalability. Across chunked variants, $c=256$ achieves the best trade-off, attaining the highest average score while maintaining comparable TTFT. Smaller chunks slightly degrade accuracy, likely due to restricted matching scope within each chunk. These results demonstrate that chunked clustering effectively amortizes computational cost with minimal impact on accuracy, and that a moderate chunk size provides the best balance between efficiency and representational capacity.

\paragraph{Compression Ratio.}
Table~\ref{tab:ablation_rate} evaluates the compression ratio scheduling defined by Equation~\ref{eq:compression_ratio}. We observe that aggressive merging with $r_{\text{init}} = 1.0$ leads to severe accuracy degradation, indicating that indiscriminate merging introduces substantial noise by aggregating low-similarity tokens. Reducing $r_{\text{init}}$ to 0.9 or 0.8 significantly improves performance, as it prioritizes higher-confidence matches. Applying a decaying schedule further enhances accuracy. As clustering proceeds, decreasing $r$ progressively enforces stricter merging criteria, effectively filtering out weaker similarities in later rounds. This leads to more reliable centroid representations across iterations. The best configuration ($r_{\text{init}} = 0.80$, $\delta_r = 0.20$) achieves the highest average score while maintaining comparable TTFT across all context lengths.

\section{Conclusion and Limitations}

This paper presents CentroidKV, a simple yet effective framework for online KV cache clustering that improves the efficiency of long-context LLM inference. CentroidKV reduces KV cache memory usage by up to 75\% without compromising accuracy on most tasks, while accelerating the decoding stage by up to $1.92\times$ and increasing the throughput by up to $4\times$. However, our work focuses on compressing the KV cache on the GPU and does not investigate memory offloading strategies. A promising direction for future research is to perform clustering on the CPU and transfer the resulting centroids to the GPU. Additionally, CentroidKV currently employs a manually specified compression-ratio schedule and restricts cache reduction to at most half at each clustering step. This could be addressed by designing an adaptive compression strategy in future work.

\newpage

\bibliography{ref}

@article{zhang2023h2o,
  title={H2o: Heavy-hitter oracle for efficient generative inference of large language models},
  author={Zhang, Zhenyu and Sheng, Ying and Zhou, Tianyi and Chen, Tianlong and Zheng, Lianmin and Cai, Ruisi and Song, Zhao and Tian, Yuandong and R{\'e}, Christopher and Barrett, Clark and others},
  journal={Advances in Neural Information Processing Systems},
  volume={36},
  pages={34661--34710},
  year={2023}
}

@article{xiao2023efficient,
  title={Efficient streaming language models with attention sinks},
  author={Xiao, Guangxuan and Tian, Yuandong and Chen, Beidi and Han, Song and Lewis, Mike},
  journal={arXiv preprint arXiv:2309.17453},
  year={2023}
}

@article{li2024snapkv,
  title={Snapkv: Llm knows what you are looking for before generation},
  author={Li, Yuhong and Huang, Yingbing and Yang, Bowen and Venkitesh, Bharat and Locatelli, Acyr and Ye, Hanchen and Cai, Tianle and Lewis, Patrick and Chen, Deming},
  journal={arXiv preprint arXiv:2404.14469},
  year={2024}
}

@article{ge2023model,
  title={Model tells you what to discard: Adaptive kv cache compression for llms},
  author={Ge, Suyu and Zhang, Yunan and Liu, Liyuan and Zhang, Minjia and Han, Jiawei and Gao, Jianfeng},
  journal={arXiv preprint arXiv:2310.01801},
  year={2023}
}

@article{hooper2024kvquant,
  title={Kvquant: Towards 10 million context length llm inference with kv cache quantization},
  author={Hooper, Coleman and Kim, Sehoon and Mohammadzadeh, Hiva and Mahoney, Michael W and Shao, Yakun Sophia and Keutzer, Kurt and Gholami, Amir},
  journal={arXiv preprint arXiv:2401.18079},
  year={2024}
}

@article{liu2024kivi,
  title={Kivi: A tuning-free asymmetric 2bit quantization for kv cache},
  author={Liu, Zirui and Yuan, Jiayi and Jin, Hongye and Zhong, Shaochen and Xu, Zhaozhuo and Braverman, Vladimir and Chen, Beidi and Hu, Xia},
  journal={arXiv preprint arXiv:2402.02750},
  year={2024}
}

@article{wan2024d2o,
  title={D2O: Dynamic Discriminative Operations for Efficient Generative Inference of Large Language Models},
  author={Wan, Zhongwei and Wu, Xinjian and Zhang, Yu and Xin, Yi and Tao, Chaofan and Zhu, Zhihong and Wang, Xin and Luo, Siqi and Xiong, Jing and Zhang, Mi},
  journal={arXiv preprint arXiv:2406.13035},
  year={2024}
}

@article{wang2024model,
  title={Model tells you where to merge: Adaptive kv cache merging for llms on long-context tasks},
  author={Wang, Zheng and Jin, Boxiao and Yu, Zhongzhi and Zhang, Minjia},
  journal={arXiv preprint arXiv:2407.08454},
  year={2024}
}

@inproceedings{zhangcam,
  title={CaM: Cache Merging for Memory-efficient LLMs Inference},
  author={Zhang, Yuxin and Du, Yuxuan and Luo, Gen and Zhong, Yunshan and Zhang, Zhenyu and Liu, Shiwei and Ji, Rongrong},
  booktitle={Forty-first International Conference on Machine Learning}
}

@article{bolya2022token,
  title={Token merging: Your vit but faster},
  author={Bolya, Daniel and Fu, Cheng-Yang and Dai, Xiaoliang and Zhang, Peizhao and Feichtenhofer, Christoph and Hoffman, Judy},
  journal={arXiv preprint arXiv:2210.09461},
  year={2022}
}

@inproceedings{bolya2023token,
  title={Token merging for fast stable diffusion},
  author={Bolya, Daniel and Hoffman, Judy},
  booktitle={Proceedings of the IEEE/CVF conference on computer vision and pattern recognition},
  pages={4599--4603},
  year={2023}
}

@inproceedings{kim2024token,
  title={Token fusion: Bridging the gap between token pruning and token merging},
  author={Kim, Minchul and Gao, Shangqian and Hsu, Yen-Chang and Shen, Yilin and Jin, Hongxia},
  booktitle={Proceedings of the IEEE/CVF Winter Conference on Applications of Computer Vision},
  pages={1383--1392},
  year={2024}
}

@article{tran2024accelerating,
  title={Accelerating Transformers with Spectrum-Preserving Token Merging},
  author={Tran, Hoai-Chau and Nguyen, Duy MH and Nguyen, Duy M and Nguyen, Trung-Tin and Le, Ngan and Xie, Pengtao and Sonntag, Daniel and Zou, James Y and Nguyen, Binh T and Niepert, Mathias},
  journal={arXiv preprint arXiv:2405.16148},
  year={2024}
}

@article{dosovitskiy2020image,
  title={An image is worth 16x16 words: Transformers for image recognition at scale},
  author={Dosovitskiy, Alexey},
  journal={arXiv preprint arXiv:2010.11929},
  year={2020}
}

@article{liu2024clusterkv,
  title={ClusterKV: Manipulating LLM KV Cache in Semantic Space for Recallable Compression},
  author={Liu, Guangda and Li, Chengwei and Zhao, Jieru and Zhang, Chenqi and Guo, Minyi},
  journal={arXiv preprint arXiv:2412.03213},
  year={2024}
}

@misc{together2023llama2,
  author = {Together},
  title = {Llama-2-7B-32K-Instruct — and fine-tuning for Llama-2 models with Together API},
  year = {2023},
  month = {June},
  url = {https://www.together.ai/blog/llama-2-7b-32k-instruct}
}

@article{xiao2024duoattention,
  title={Duoattention: Efficient long-context llm inference with retrieval and streaming heads},
  author={Xiao, Guangxuan and Tang, Jiaming and Zuo, Jingwei and Guo, Junxian and Yang, Shang and Tang, Haotian and Fu, Yao and Han, Song},
  journal={arXiv preprint arXiv:2410.10819},
  year={2024}
}

@article{feng2024ada,
  title={Ada-kv: Optimizing kv cache eviction by adaptive budget allocation for efficient llm inference},
  author={Feng, Yuan and Lv, Junlin and Cao, Yukun and Xie, Xike and Zhou, S Kevin},
  journal={arXiv preprint arXiv:2407.11550},
  year={2024}
}

@article{tang2024razorattention,
  title={Razorattention: Efficient kv cache compression through retrieval heads},
  author={Tang, Hanlin and Lin, Yang and Lin, Jing and Han, Qingsen and Hong, Shikuan and Yao, Yiwu and Wang, Gongyi},
  journal={arXiv preprint arXiv:2407.15891},
  year={2024}
}

@article{cai2024pyramidkv,
  title={Pyramidkv: Dynamic kv cache compression based on pyramidal information funneling},
  author={Cai, Zefan and Zhang, Yichi and Gao, Bofei and Liu, Yuliang and Liu, Tianyu and Lu, Keming and Xiong, Wayne and Dong, Yue and Chang, Baobao and Hu, Junjie and others},
  journal={arXiv preprint arXiv:2406.02069},
  year={2024}
}

@article{yang2024pyramidinfer,
  title={PyramidInfer: Pyramid KV Cache Compression for High-throughput LLM Inference},
  author={Yang, Dongjie and Han, XiaoDong and Gao, Yan and Hu, Yao and Zhang, Shilin and Zhao, Hai},
  journal={arXiv preprint arXiv:2405.12532},
  year={2024}
}

@article{liu2024scissorhands,
  title={Scissorhands: Exploiting the persistence of importance hypothesis for llm kv cache compression at test time},
  author={Liu, Zichang and Desai, Aditya and Liao, Fangshuo and Wang, Weitao and Xie, Victor and Xu, Zhaozhuo and Kyrillidis, Anastasios and Shrivastava, Anshumali},
  journal={Advances in Neural Information Processing Systems},
  volume={36},
  year={2024}
}

@article{achiam2023gpt,
  title={Gpt-4 technical report},
  author={Achiam, Josh and Adler, Steven and Agarwal, Sandhini and Ahmad, Lama and Akkaya, Ilge and Aleman, Florencia Leoni and Almeida, Diogo and Altenschmidt, Janko and Altman, Sam and Anadkat, Shyamal and others},
  journal={arXiv preprint arXiv:2303.08774},
  year={2023}
}

@article{team2024gemini,
  title={Gemini 1.5: Unlocking multimodal understanding across millions of tokens of context},
  author={Team, Gemini and Georgiev, Petko and Lei, Ving Ian and Burnell, Ryan and Bai, Libin and Gulati, Anmol and Tanzer, Garrett and Vincent, Damien and Pan, Zhufeng and Wang, Shibo and others},
  journal={arXiv preprint arXiv:2403.05530},
  year={2024}
}

@article{touvron2023llama,
  title={Llama: Open and efficient foundation language models},
  author={Touvron, Hugo and Lavril, Thibaut and Izacard, Gautier and Martinet, Xavier and Lachaux, Marie-Anne and Lacroix, Timoth{\'e}e and Rozi{\`e}re, Baptiste and Goyal, Naman and Hambro, Eric and Azhar, Faisal and others},
  journal={arXiv preprint arXiv:2302.13971},
  year={2023}
}

@article{dao2022flashattention,
  title={Flashattention: Fast and memory-efficient exact attention with io-awareness},
  author={Dao, Tri and Fu, Dan and Ermon, Stefano and Rudra, Atri and R{\'e}, Christopher},
  journal={Advances in Neural Information Processing Systems},
  volume={35},
  pages={16344--16359},
  year={2022}
}

@article{bai2023longbench,
  title={Longbench: A bilingual, multitask benchmark for long context understanding},
  author={Bai, Yushi and Lv, Xin and Zhang, Jiajie and Lyu, Hongchang and Tang, Jiankai and Huang, Zhidian and Du, Zhengxiao and Liu, Xiao and Zeng, Aohan and Hou, Lei and others},
  journal={arXiv preprint arXiv:2308.14508},
  year={2023}
}

@article{wolf2019huggingface,
  title={Huggingface's transformers: State-of-the-art natural language processing},
  author={Wolf, T},
  journal={arXiv preprint arXiv:1910.03771},
  year={2019}
}

@article{merity2016pointer,
  title={Pointer sentinel mixture models},
  author={Merity, Stephen and Xiong, Caiming and Bradbury, James and Socher, Richard},
  journal={arXiv preprint arXiv:1609.07843},
  year={2016}
}

@article{meta2024introducing,
  title={Introducing meta llama 3: The most capable openly available llm to date},
  author={Meta, AI},
  journal={Meta AI},
  year={2024}
}

@inproceedings{rombach2022high,
  title={High-resolution image synthesis with latent diffusion models},
  author={Rombach, Robin and Blattmann, Andreas and Lorenz, Dominik and Esser, Patrick and Ommer, Bj{\"o}rn},
  booktitle={Proceedings of the IEEE/CVF conference on computer vision and pattern recognition},
  pages={10684--10695},
  year={2022}
}

@article{xu2024think,
  title={Think: Thinner key cache by query-driven pruning},
  author={Xu, Yuhui and Jie, Zhanming and Dong, Hanze and Wang, Lei and Lu, Xudong and Zhou, Aojun and Saha, Amrita and Xiong, Caiming and Sahoo, Doyen},
  journal={arXiv preprint arXiv:2407.21018},
  year={2024}
}

@article{shi2024discovering,
  title={Discovering the gems in early layers: Accelerating long-context llms with 1000x input token reduction},
  author={Shi, Zhenmei and Ming, Yifei and Nguyen, Xuan-Phi and Liang, Yingyu and Joty, Shafiq},
  journal={arXiv preprint arXiv:2409.17422},
  year={2024}
}

@misc{kamradt2024needle,
  title={Needle In A Haystack: Evaluating Long-Context Retrieval in Language Models},
  author={Kamradt, Greg},
  year={2024},
  howpublished={Online},
  url={https://github.com/gkamradt/LLMTest_NeedleInAHaystack},
  note={A benchmark for testing LLMs' ability to retrieve specific information from long-context inputs.}
}

@article{hsieh2024ruler,
  title={RULER: What's the Real Context Size of Your Long-Context Language Models?},
  author={Cheng-Ping Hsieh and Simeng Sun and Samuel Kriman and Shantanu Acharya and Dima Rekesh and Fei Jia and Yang Zhang and Boris Ginsburg},
  year={2024},
  journal={arXiv preprint arXiv:2404.06654},
}

@article{devoto2025expectedattention,
  title={Expected Attention: KV Cache Compression by Estimating Attention from Future Queries Distribution},
  author={Devoto, Alessio and Jeblick, Maximilian and J{\'e}gou, Simon},
  journal={arXiv preprint arXiv:2510.00636},
  year={2025},
  url={https://arxiv.org/abs/2510.00636}
}

@misc{jiang2023mistral7b,
      title={Mistral 7B}, 
      author={Albert Q. Jiang and Alexandre Sablayrolles and Arthur Mensch and Chris Bamford and Devendra Singh Chaplot and Diego de las Casas and Florian Bressand and Gianna Lengyel and Guillaume Lample and Lucile Saulnier and Lélio Renard Lavaud and Marie-Anne Lachaux and Pierre Stock and Teven Le Scao and Thibaut Lavril and Thomas Wang and Timothée Lacroix and William El Sayed},
      year={2023},
      eprint={2310.06825},
      archivePrefix={arXiv},
      primaryClass={cs.CL},
      url={https://arxiv.org/abs/2310.06825}, 
}

@inproceedings{kwon2023efficient,
  title={Efficient memory management for large language model serving with pagedattention},
  author={Kwon, Woosuk and Li, Zhuohan and Zhuang, Siyuan and Sheng, Ying and Zheng, Lianmin and Yu, Cody Hao and Gonzalez, Joseph and Zhang, Hao and Stoica, Ion},
  booktitle={Proceedings of the 29th symposium on operating systems principles},
  pages={611--626},
  year={2023}
}

@book{efron1994introduction,
  title={An introduction to the bootstrap},
  author={Efron, Bradley and Tibshirani, Robert J},
  year={1994},
  publisher={Chapman and Hall/CRC}
}
\bibliographystyle{tmlr}

\appendix
\section{Theoretical Results and Proof}
\label{app:theory}

We theoretically justify the optimality of our partitioning strategy which divides each chunk into two sets $A$ and $B$ in an alternating manner. Intuitively, the partitioning strategy should retain the edges with the highest similarities to yield better clusters. Based on Observation 2 in Section~\ref{sec:observation}, we consider a convex and monotonically decreasing score function $f:\mathbb{N}\rightarrow\mathbb{R}$ that maps the distance to an importance score. With the score function $f$, the optimal partitioning can be achieved by solving the following optimization problem:
\begin{align}
    \max_{A,B}\quad\sum_{x\in A}\sum_{y\in B}f(|x-y|).\label{eq:opt-obj}
\end{align}

The following theorem states that, for any score function $f$ that is convex and monotonically decreasing, the alternating partitioning strategy we propose is always the optimal solution of problem Equation~(\ref{eq:opt-obj}).
\begin{theorem}\label{thm:opt-pattern}
    Define partition set $\mathcal{P}_{2n}=\{(A,B)\mid |A|=|B|=n,\ \mathrm{and}\ A\cup B=[2n]\}$. If function $f:[2n-1]\rightarrow\mathbb{R}$ satisfies $f(1)-f(2)\ge f(2)-f(3)\ge\cdots\ge f(2n-2)-f(2n-1)\ge0$, it holds that 
    \begin{align*}
        (A_0,B_0)=&(\{1,3,\cdots,2n-1\},\{2,4,\cdots,2n\})\\
        \in&\mathop{\arg\max}_{(A,B)\in\mathcal{P}_{2n}}\sum_{x\in A}\sum_{y\in B}f(|x-y|).
    \end{align*}
    Here, we use notation $[k]$ to denote the set of positive integers no larger than $k$, \textit{i.e.}, $[k]:=[1,k]\cap\mathbb{Z}$.
\end{theorem}

\begin{proof} 
    When $n=1$, the result is trivial. In the following, we assume $n\ge2$. Consider the following mapping:
    \begin{align*}
        \phi_{2n}:\mathbb{Z}\rightarrow&[2n]\\
        x\mapsto&\begin{cases}1,&x<1;\\
        x,&x\in[2n];\\
        2n,&x>2n;
        \end{cases}
    \end{align*}
    For any $l\in[2n-1]$, define $\mathcal{S}_{2n,l}:=\{(\phi_{2n}(x),\phi_{2n}(x+l))\mid x\in\mathbb{Z}\cap[-l+2,2n-1]\}$. For $\forall (A,B)\in\mathcal{P}_{2n}$, define 
    \begin{align}
        \mathcal{E}_{A,B}:=&\{(\min\{a,b\},\max\{a,b\})\mid a\in A\ \text{and}\ b\in B\},\\
        \mathcal{D}_{A,B,l}:=&\{(a,b)\in\mathcal{E}_{A,B}\mid b-a=l\},\\
        \mathcal{T}_{A,B,l}:=&\{(a,b;c,d)\in[2n]^4\mid (a,b)\in\mathcal{S}_{2n,l},\ (c,d)\in\mathcal{E}_{A,B},\ \text{and}\ [c,d]\subseteq[a,b]\},
    \end{align}
    Now we calculate the number of elements in $\mathcal{T}_{A,B,l}$, \textit{i.e.}, $|\mathcal{T}_{A,B,l}|$. Define $\mathcal{K}_{A,B,a,b}:=\{(c,d)\in\mathcal{E}_{A,B}\mid[c,d]\subseteq[a,b]\}$, it holds that
    \begin{align}
        |\mathcal{T}_{A,B,l}|=\sum_{(a,b)\in\mathcal{S}_{2n,l}}|\mathcal{K}_{A,B,a,b}|.\label{eq:pfthm-tabl}
    \end{align}
    Note that
    \begin{align}
        |\mathcal{K}_{A,B,a,b}|=|[a,b]\cap A|\cdot |[a,b]\cap B|\le \left\lfloor\frac{b-a+1}{2}\right\rfloor\cdot\left\lceil\frac{b-a+1}{2}\right\rceil,\label{eq:pfthm-kab}
    \end{align}
    applying \eqref{eq:pfthm-kab} to \eqref{eq:pfthm-tabl} yields
    \begin{align}
        |\mathcal{T}_{A,B,l}|\le&\sum_{(a,b)\in\mathcal{S}_{2n,l}}\left\lfloor\frac{b-a+1}{2}\right\rfloor\cdot\left\lceil\frac{b-a+1}{2}\right\rceil\nonumber\\
        =&\sum_{i=2}^l\left\lfloor\frac{i-1+1}{2}\right\rfloor\cdot\left\lceil\frac{i-1+1}{2}\right\rceil+\sum_{i=1}^{2n-l}\left\lfloor\frac{(i+l)-i+1}{2}\right\rfloor\cdot\left\lceil\frac{(i+l)-i+1}{2}\right\rceil+\nonumber\\
        &+\sum_{i=2n-l+1}^{2n-1}\left\lfloor\frac{2n-i+1}{2}\right\rfloor\cdot\left\lceil\frac{2n-i+1}{2}\right\rceil\nonumber\\
        =&\begin{cases} -\frac{1}{12}l^3+\frac{2n-1}{4}l^2+\frac{6n-1}{6}l,&2\mid l;\\
        -\frac{1}{12}l^3+\frac{2n-1}{4}l^2+\frac{12n-5}{12}l+\frac{2n-1}{4},&2\nmid l.
        \end{cases}\triangleq c_{2n,l}.\label{eq:pfthm-tabl2}
    \end{align}
    On the other hand, define $\mathcal{K}'_{2n,l,c,d}:=\{(a,b)\in\mathcal{S}_{2n,l}\mid [c,d]\subseteq[a,b]\}$, we have 
    \begin{align*}
        |\mathcal{K}'_{2n,l,c,d}|=\max(l+1-d+c,0),
    \end{align*}
    thus 
    \begin{align}
        |\mathcal{T}_{A,B,l}|=&\sum_{(c,d)\in\mathcal{E}_{A,B}}|\mathcal{K}'_{2n,l,c,d}|\nonumber\\
        =&\sum_{(c,d)\in\mathcal{E}_{A,B}}\max(l+1-d+c,0)\nonumber\\
        =&\sum_{i=1}^l(l+1-i)|\mathcal{D}_{A,B,i}|.\label{eq:pfthm-tabl3}
    \end{align}
    Combining \eqref{eq:pfthm-tabl2}\eqref{eq:pfthm-tabl3} yields
    \begin{align}
        \sum_{i=1}^l(l+1-i)|\mathcal{D}_{A,B,i}|\le c_{2n,l}.\label{eq:pfthm-dabi}
    \end{align}
    Define $a_i=f(i)-f(i+1)$ for $i\in[2n-2]$, $b_i=a_i-a_{i+1}$ for $i\in[2n-3]$, and let $b_{2n-2}=a_{2n-2}$, it holds that $b_1,b_2,\cdots,b_{2n-2}\ge0$. Note that
    \begin{align*}
    f(i)=&f(2n-1)+\sum_{j=i}^{2n-2}a_j,\\
    =&f(2n-1)+\sum_{j=i}^{2n-2}\sum_{k=j}^{2n-2}b_k,\\
    =&f(2n-1)+\sum_{j=i}^{2n-2}(j-i+1)b_j,\quad\forall i\in[2n-2],
    \end{align*}
    we have
    \begin{align}
        \sum_{x\in A}\sum_{y\in B}f(|x-y|)=&\sum_{i=1}^{2n-1}|\mathcal{D}_{A,B,i}|f(i)\nonumber\\
        =&\sum_{i=1}^{2n-1}|\mathcal{D}_{A,B,i}|\left(f(2n-1)+\sum_{j=i}^{2n-2}(j-i+1)b_j\right)\nonumber\\
        =&|\mathcal{E}_{A,B}|f(2n-1)+\sum_{j=1}^{2n-2}\sum_{i=1}^j(j+1-i)|\mathcal{D}_{A,B,i}|b_j\nonumber\\
        \le&n^2f(2n-1)+\sum_{j=1}^{2n-2}c_{2n,j}b_j,\label{eq:pfthm-sfxy0}
    \end{align}
    where the last inequality uses $|\mathcal{E}_{A,B}|=n^2$ and \eqref{eq:pfthm-dabi}. If $A_0=\{1,3,\cdots,2n-1\}$ and $B_0=\{2,4,\cdots,2n\}$, we have
    \begin{align}
        \sum_{x\in A_0}\sum_{y\in B_0}f(|x-y|)=&\sum_{i=1}^{2n-1}|\mathcal{D}_{A_0,B_0,i}|f(i)=\sum_{i=1}^n|\mathcal{D}_{A_0,B_0,2i-1}|f(2i-1)\nonumber\\
        =&\sum_{i=1}^n(2n-2i+1)\left(f(2n-1)+\sum_{j=2i-1}^{2n-2}(j-2i+2)b_j\right)\nonumber\\
        =&n^2f(2n-1)+\sum_{j=1}^{2n-2}\sum_{i=1}^{\left\lfloor\frac{j+1}{2}\right\rfloor}(2n-2i+1)(j-2i+2)b_j.\label{eq:pfthm-sfxy}
    \end{align}
    Note that 
    \begin{align}
        \sum_{i=1}^{\left\lfloor\frac{j+1}{2}\right\rfloor}(2n-2i+1)(j-2i+2)=\begin{cases}-\frac{1}{12}l^3+\frac{2n-1}{4}l^2+\frac{6n-1}{6}l,&2\mid l;\\
        -\frac{1}{12}l^3+\frac{2n-1}{4}l^2+\frac{12n-5}{12}l+\frac{2n-1}{4},&2\nmid l.
        \end{cases}=c_{2n,j},\label{eq:pfthm-c2nj}
    \end{align}
    combining \eqref{eq:pfthm-sfxy}\eqref{eq:pfthm-c2nj} yields
    \begin{align}
        \sum_{x\in A_0}\sum_{y\in B_0}f(|x-y|)=n^2f(2n-1)+\sum_{j=1}^{2n-2}c_{2n,j}b_j.\label{eq:pfthm-sfxy2}
    \end{align}
    Combining \eqref{eq:pfthm-sfxy0}\eqref{eq:pfthm-sfxy2}, we obtain
    \begin{align*}(A_0,B_0)\in\mathop{\arg\max}_{(A,B)\in\mathcal{P}_{2n}}\sum_{x\in A}\sum_{y\in B}f(|x-y|),
    \end{align*}
    which concludes the proof.
\end{proof}

\section{Experimental Setups}
\label{app:exp_set}
We compare CentroidKV with other compression-based baselines based on KVPress \cite{devoto2025expectedattention}, which implements multiple KV cache compression methods and benchmarks. All baseline methods are evaluated using their recommended settings. For StreamingLLM \citep{xiao2023efficient}, the number of initial sink tokens is 4. For SnapKV \citep{li2024snapkv}, window size is 64 and pooling kernel size is 5. For PyramidKV \citep{cai2024pyramidkv}, the hyperparameter which controls the pyramid's shape and steepness is set to 20.

Parameters of CentroidKV maintain the same across the entire evaluation process, which are reported in the implementation details of Section~\ref{sec:experiment}. Ablation studies on chunk size and the compression ratio (including $r_{\text{init}}$ and $\delta_r$) are demonstrated in Section~\ref{sec:experiment}. For number of attention sink and recent tokens, we simply follow the recommended settings in DuoAttention \cite{xiao2024duoattention}.

\begin{table}[t]
\caption{Detailed results of LongBench datasets with 25\% KV cache budget on Llama-3.1-8B-Instruct.}
\label{tab:longbench_llama_25pct}
\begin{center}
\begin{small}
\begin{tabular}{lccccr}
\toprule
Dataset & StreamingLLM & SnapKV & PyramidKV & CentroidKV & Full \\
\midrule
2WikiMQA & 27.92 & \textbf{42.44} & 39.83 & 33.18 & 51.33 \\
GovReport & 28.79 & \textbf{29.24} & 28.23 & 27.24 & 35.19 \\
HotpotQA & 41.95 & \textbf{55.18} & 55.13 & 52.08 & 59.80 \\
LCC & 50.84 & \textbf{53.83} & 53.61 & 51.21 & 53.31 \\
MultiNews & \textbf{23.91} & 23.35 & 23.41 & 23.10 & 26.99 \\
MultiFieldQA-en & 25.62 & 35.01 & 34.79 & \textbf{39.81} & 56.31 \\
Musique & 20.07 & \textbf{28.18} & 25.19 & 25.58 & 33.51 \\
NarrativeQA & 23.25 & \textbf{28.43} & 26.92 & 27.03 & 30.99 \\
Passage Count & 7.00 & 9.10 & 9.55 & \textbf{10.00} & 10.70 \\
PassageRetrieval-en & 33.50 & 90.00 & \textbf{93.50} & 92.00 & 100.00 \\
Qasper & 24.38 & 31.36 & 30.08 & \textbf{33.27} & 47.39 \\
QMSum & 20.69 & 22.41 & 22.20 & \textbf{23.42} & 25.34 \\
RepoBench-P & \textbf{50.50} & 47.72 & 47.28 & 49.30 & 47.23 \\
SAMSum & 35.70 & 41.43 & \textbf{42.26} & 41.74 & 40.77 \\
TREC & 31.50 & \textbf{37.00} & 34.00 & 33.00 & 29.00 \\
TriviaQA & \textbf{92.12} & 91.54 & 91.73 & 91.14 & 91.71 \\
\bottomrule
\end{tabular}
\end{small}
\end{center}
\end{table}

\begin{table}[t]
\caption{Detailed results of LongBench datasets with 25\% KV cache budget on Mistral-7B-Instruct-v0.2.}
\label{tab:longbench_mistral_25pct}
\begin{center}
\begin{small}
\begin{tabular}{lccccr}
\toprule
Dataset & StreamingLLM & SnapKV & PyramidKV & CentroidKV & Full \\
\midrule
2WikiMQA & \textbf{15.83} & 13.75 & 13.59 & 15.62 & 20.97 \\
GovReport & \textbf{28.38} & 27.27 & 26.20 & 26.87 & 32.34 \\
HotpotQA & 21.59 & 23.96 & 22.98 & \textbf{25.98} & 35.34 \\
LCC & 48.37 & \textbf{51.25} & 50.79 & 47.34 & 51.35 \\
MultiNews & 22.87 & 23.16 & 23.23 & \textbf{23.60} & 26.55 \\
MultiFieldQA-en & 22.46 & 31.65 & 31.27 & \textbf{32.20} & 45.94 \\
Musique & 10.42 & \textbf{11.32} & 10.03 & 9.68 & 17.30 \\
NarrativeQA & 21.36 & 18.62 & 17.30 & \textbf{21.42} & 23.86 \\
Passage Count & 3.10 & \textbf{3.91} & 3.39 & 2.94 & 2.48 \\
PassageRetrieval-en & 18.48 & \textbf{68.81} & 66.77 & 35.32 & 73.76 \\
Qasper & 13.32 & 14.36 & 13.57 & \textbf{15.45} & 29.20 \\
QMSum & 20.58 & 21.59 & 21.41 & \textbf{22.94} & 24.37 \\
RepoBench-P & 47.55 & \textbf{51.08} & 50.74 & 50.09 & 51.16 \\
SAMSum & 36.12 & 38.68 & \textbf{39.10} & 36.63 & 39.61 \\
TREC & \textbf{47.50} & 41.75 & 39.25 & 23.50 & 51.25 \\
TriviaQA & 48.69 & \textbf{79.43} & 77.46 & 67.37 & 74.68 \\
\bottomrule
\end{tabular}
\end{small}
\end{center}
\end{table}

\section{Additional Results of LongBench}
\label{app:longbench_detail}

Below is the overview of datasets categorized by tasks in LongBench benchmark.
\begin{itemize}
    \item Single-document QA: NarrativeQA, Qasper, MultiFieldQA-en.
    \item Multi-document QA: HotpotQA, 2WikiMQA, Musique.
    \item Summarization: GovReport, QMSum, MultiNews.
    \item Few-shot learning: SAMSum, TREC, TriviaQA.
    \item Synthetic Tasks: PassageRetrieval-en, Passage Count.
    \item Code Completion: LCC, RepoBench-P.
\end{itemize}

Table~\ref{tab:longbench_llama_25pct} and ~\ref{tab:longbench_mistral_25pct} present the detailed results of different methods across 16 LongBench datasets on Llama and Mistral respectively. CentroidKV outperforms on semantic aggregation and reasoning-heavy tasks, while showing limitations on retrieval-intensive tasks requiring precise token preservation.

\begin{table}[t]
\caption{Efficiency evaluation within vLLM on Llama-3.1-8B-Instruct, 25\% KV cache budget of CentroidKV.}
\label{tab:vllm_llama_tpot_throughput}
\begin{center}
\small
\begin{adjustbox}{max width=\textwidth}
\begin{tabular}{ccccccc}
\toprule
Context & \multicolumn{3}{c}{TPOT (ms)}  & \multicolumn{3}{c}{Throughput (tok/s)} \\
\cmidrule(lr){2-4} \cmidrule(lr){5-7}
 Length & Full & CentroidKV &  Speedup & Full & CentroidKV & Gain \\
\midrule
4k & 13.91 & 13.65 & $1.02\times$ & 1263.8 & 2960.2 & $2.34\times$ \\
8k & 14.27 & 13.59 & $1.05\times$ & 669.3 & 2255.2 & $3.37\times$ \\
16k & 15.05 & 13.81 & $1.09\times$ & 388.1 & 1335.6 & $3.44\times$ \\
32k & 16.73 & 14.47 & $1.16\times$ & 203.7 & 664.7 & $3.26\times$ \\
64k & 20.47 & 14.93 & $1.37\times$ & 100.5 & 401.7 & $4.00\times$ \\
128k & 26.79 & 16.43 & $1.63\times$ & 50.4 & 155.2 & $3.08\times$ \\
\bottomrule
\end{tabular}
\end{adjustbox}
\end{center}
\end{table}

\section{Efficiency evaluation based on vLLM}
\label{app:vllm}

To validate that CentroidKV's compression benefits persist under realistic serving conditions with memory management overhead, we conduct experiments within vLLM \cite{kwon2023efficient}, a widely deployed production inference engine that employs PagedAttention for dynamic KV cache memory management. 

We integrate CentroidKV into vLLM's inference pipeline as a post-prefill compression step: after the prefill phase populates the paged KV cache, CentroidKV reads the key-value tensors, applies compression and writes the compressed representations back into the page table. Freed memory blocks are returned to the block manager, enabling the engine to serve additional concurrent requests. All experiments use Llama-3.1-8B-Instruct on a single A100 GPU. We evaluate across context lengths from 4K to 128K tokens and report two metrics: (1)~Time Per Output Token (TPOT) at batch size~1, measuring single-request decode latency; (2)~Peak generation throughput, measuring the maximum number of concurrent requests that fit within GPU memory.

As shown in Table~\ref{tab:vllm_llama_tpot_throughput}, CentroidKV achieves decoding speedups ranging from $1.02\times$ at 4K context to $1.63\times$ at 128K context. This confirms that the reduced KV cache footprint directly translates into lower memory-bandwidth pressure during attention computation, with the benefit growing as context length increases and attention becomes increasingly memory-bound. The throughput improvements are substantially larger. By freeing 75\% of KV cache blocks after compression, CentroidKV enables the engine to sustain 2.3--4.0$\times$ more concurrent requests within the same GPU memory.

\begin{table}[t]
\caption{CentroidKV vs.\ full KV on Llama-3.1-8B-Instruct with uniform KV cache budget 2048, evaluated in KVPress codebase.}
\footnotesize
\label{tab:eff_centroidkv_vs_full}
\centering
\begin{small}
\begin{adjustbox}{max width=\textwidth}
\begin{tabular}{clccccc}
   \toprule
    Context Length & Method & TTFT (s) & TPOT (ms) & Speedup & Memory (GB) & Mem Saving \\
    \midrule
    \multirow{2}{*}{4k}
        & Full & 0.371 & 35.53 & \multirow{2}{*}{$\times$} & 16.44 & \multirow{2}{*}{\checkmark} \\
        & CentroidKV & 0.444 & 39.23 & & 16.20 & \\
    \midrule
    \multirow{2}{*}{8k}
        & Full & 0.780 & 35.82 & \multirow{2}{*}{$\times$} & 17.92 & \multirow{2}{*}{\checkmark} \\
        & CentroidKV & 0.963 & 37.80 & & 17.19 & \\
    \midrule
    \multirow{2}{*}{16k}
        & Full & 1.735 & 38.47 & \multirow{2}{*}{\checkmark} & 20.87 & \multirow{2}{*}{\checkmark} \\
        & CentroidKV & 2.036 & \textbf{36.58} & & 19.24 & \\
    \midrule
    \multirow{2}{*}{32k}
        & Full & 4.186 & 44.67 & \multirow{2}{*}{\checkmark} & 26.76 & \multirow{2}{*}{\checkmark} \\
        & CentroidKV & 4.513 & \textbf{35.34} & & 23.49 & \\
    \midrule
    \multirow{2}{*}{64k}
        & Full & 11.300 & 59.02 & \multirow{2}{*}{\checkmark} & 38.56 & \multirow{2}{*}{\checkmark} \\
        & CentroidKV & 11.881 & \textbf{35.44} & & 32.01 & \\
   \bottomrule
\end{tabular}
\end{adjustbox}
\end{small}
\vskip 0.1in
\end{table}

\begin{table}[t]
\caption{ClusterKV vs.\ full KV on Llama-3.1-8B-Instruct with uniform KV cache budget 2048, evaluated in ClusterKV's official codebase with their customized kernels.}
\footnotesize
\label{tab:eff_clusterkv_official_vs_full}
\centering
\begin{small}
\begin{adjustbox}{max width=\textwidth}
\begin{tabular}{clccccc}
   \toprule
    Context Length & Method & TTFT (s) & TPOT (ms) & Speedup & Memory (GB) & Mem Saving \\
    \midrule
    \multirow{2}{*}{4k}
        & Full & 0.355 & 13.89 & \multirow{2}{*}{$\times$} & 18.46 & \multirow{2}{*}{$\times$} \\
        & ClusterKV & 0.453 & 16.44 & & 18.50 & \\
    \midrule
    \multirow{2}{*}{8k}
        & Full & 0.758 & 14.01 & \multirow{2}{*}{$\times$} & 21.84 & \multirow{2}{*}{$\times$} \\
        & ClusterKV & 0.962 & 16.45 & & 21.89 & \\
    \midrule
    \multirow{2}{*}{16k}
        & Full & 1.683 & 14.92 & \multirow{2}{*}{$\times$} & 28.63 & \multirow{2}{*}{$\times$} \\
        & ClusterKV & 2.084 & 16.65 & & 28.69 & \\
    \midrule
    \multirow{2}{*}{32k}
        & Full & 4.179 & 16.30 & \multirow{2}{*}{$\times$} & 42.16 & \multirow{2}{*}{$\times$} \\
        & ClusterKV & 5.134 & 17.05 & & 42.25 & \\
   \bottomrule
\end{tabular}
\end{adjustbox}
\end{small}
\vskip 0.1in
\end{table}

\section{Comparison with ClusterKV}
\label{app:clusterkv}

ClusterKV~\citep{liu2024clusterkv} and CentroidKV both use clustering but for undamentally different design objectives. ClusterKV uses clustering as an recall mechanism while retaining the full KV cache, whereas CentroidKV directly replaces raw KV states with cluster centroids as a compressed representation. As a result, ClusterKV prioritizes information preservation and can be viewed as an accuracy-oriented retrieval baseline, while CentroidKV explicitly targets KV memory reduction and decoding efficiency.

To make this distinction explicit, we evaluate ClusterKV using its official implementation (with custom CUDA kernels) and compare both methods against their respective dense-KV baselines on Llama-3.1-8B-Instruct. Results are reported in Table~\ref{tab:eff_centroidkv_vs_full} and Table~\ref{tab:eff_clusterkv_official_vs_full}. All experiments use FP16 precision, a fixed KV cache budget of 2048, and a decoding length of 256 across varying context lengths. We exclude the 64k setting for ClusterKV due to infeasibility.

Since the two methods rely on different system implementations, we focus on their relative overhead with respect to their own full-KV baselines. Under this protocol, ClusterKV consistently introduces additional overhead without improving memory efficiency. As shown in Table~\ref{tab:eff_clusterkv_official_vs_full}, TTFT increases by 22.9\%--27.4\% across 4k to 32k contexts, while TPOT degrades by 4.6\%--18.4\%. Moreover, end-to-end GPU memory usage remains effectively unchanged (or slightly higher), indicating that ClusterKV does not achieve actual KV footprint reduction.

In contrast, CentroidKV exhibits a different efficiency profile. As shown in Table~\ref{tab:eff_centroidkv_vs_full}, it incurs a moderate increase in TTFT due to the compression cost, but this overhead does not scale with context length. More importantly, CentroidKV significantly improves decoding efficiency at long contexts: TPOT becomes lower than full KV starting from 16k, yielding substantial per-token speedups. In addition, it achieves consistent memory savings that grow with context length, reducing GPU memory by 1.5\% at 4k and up to 17.0\% at 64k.

Overall, while both methods exploit similarity in the key space, their system-level trade-offs differ substantially. CentroidKV translates redundancy into compact in-GPU representations, leading to tangible gains in both memory and long-context decoding efficiency. In contrast, ClusterKV maintains a recallable clustered index on top of the full KV cache, where the additional indexing, selection, and data movement introduce persistent overhead without reducing the underlying memory footprint.

\begin{table}[t]
\caption{Bootstrap summary on RULER for Llama-3.1-8B-Instruct. Values are reported as mean $\pm$ std.}
\footnotesize
\label{tab:bootstrap-ruler-llama31-8b}
\centering
\begin{small}
\setlength{\tabcolsep}{5pt}
\begin{tabular}{lccc}
\toprule
Method & 75\% & 50\% & 25\% \\
\midrule
StreamingLLM & 80.69 $\pm$ 0.96 & 58.59 $\pm$ 1.16 & 37.20 $\pm$ 1.02 \\
SnapKV & 80.97 $\pm$ 0.80 & 68.67 $\pm$ 0.90 & 41.78 $\pm$ 0.94 \\
PyramidKV & 81.95 $\pm$ 0.77 & 68.33 $\pm$ 0.90 & 41.59 $\pm$ 0.93 \\
CentroidKV & \textbf{90.66 $\pm$ 0.70} & \textbf{79.24 $\pm$ 0.77} & \textbf{60.26 $\pm$ 0.78} \\
\bottomrule
\end{tabular}
\end{small}
\vskip 0.1in
\end{table}

\begin{table}[t]
\caption{Bootstrap summary on RULER for Mistral-7B-Instruct-v0.2. Values are reported as mean $\pm$ std.}
\footnotesize
\label{tab:bootstrap-ruler-mistral7b}
\centering
\begin{small}
\setlength{\tabcolsep}{5pt}
\begin{tabular}{lccc}
\toprule
Method & 75\% & 50\% & 25\% \\
\midrule
StreamingLLM & 75.42 $\pm$ 1.01 & 56.57 $\pm$ 1.13 & 37.64 $\pm$ 1.04 \\
SnapKV & 58.38 $\pm$ 1.01 & 42.06 $\pm$ 0.99 & 32.77 $\pm$ 0.89 \\
PyramidKV & 62.21 $\pm$ 1.01 & 39.81 $\pm$ 0.95 & 30.58 $\pm$ 0.90 \\
CentroidKV & \textbf{82.60 $\pm$ 0.84} & \textbf{58.81 $\pm$ 0.99} & \textbf{38.02 $\pm$ 0.78} \\
\bottomrule
\end{tabular}
\end{small}
\vskip 0.1in
\end{table}

\begin{table}[t]
\caption{Bootstrap summary on LongBench for Llama-3.1-8B-Instruct. Values are reported as mean $\pm$ std.}
\footnotesize
\label{tab:bootstrap-longbench-llama31-8b}
\centering
\begin{small}
\setlength{\tabcolsep}{5pt}
\begin{tabular}{lccc}
\toprule
Method & 75\% & 50\% & 25\% \\
\midrule
StreamingLLM & 41.10 $\pm$ 0.56 & 37.69 $\pm$ 0.56 & 33.61 $\pm$ 0.52 \\
SnapKV & 45.86 $\pm$ 0.53 & 44.54 $\pm$ 0.53 & \textbf{41.64 $\pm$ 0.54} \\
PyramidKV & 45.82 $\pm$ 0.53 & 44.52 $\pm$ 0.52 & 41.11 $\pm$ 0.52 \\
CentroidKV & \textbf{46.77 $\pm$ 0.54} & \textbf{45.39 $\pm$ 0.53} & 40.82 $\pm$ 0.53 \\
\bottomrule
\end{tabular}
\end{small}
\end{table}

\begin{table}[t]
\caption{Bootstrap summary on LongBench for Mistral-7B-Instruct-v0.2. Values are reported as mean $\pm$ std.}
\footnotesize
\label{tab:bootstrap-longbench-mistral7b}
\centering
\begin{small}
\setlength{\tabcolsep}{5pt}
\begin{tabular}{lccc}
\toprule
Method & 75\% & 50\% & 25\% \\
\midrule
StreamingLLM & 33.50 $\pm$ 0.49 & 30.30 $\pm$ 0.48 & 26.66 $\pm$ 0.44 \\
SnapKV & \textbf{36.98 $\pm$ 0.47} & \textbf{35.64 $\pm$ 0.46} & \textbf{32.55 $\pm$ 0.45} \\
PyramidKV & 36.65 $\pm$ 0.48 & 33.99 $\pm$ 0.45 & 31.70 $\pm$ 0.44 \\
CentroidKV & 36.42 $\pm$ 0.49 & 34.35 $\pm$ 0.48 & 28.56 $\pm$ 0.45 \\
\bottomrule
\end{tabular}
\end{small}
\end{table}

\section{Statistical Significance Analysis}

Given that aggregate scores across benchmarks sometimes differ by only a few percentage points, we conduct bootstrap hypothesis testing to verify that the observed performance differences are statistically meaningful rather than artifacts of evaluation variance. We employ the non-parametric bootstrap~\citep{efron1994introduction} to estimate the sampling distribution of aggregate scores. We report bootstrap standard deviations and 95\% percentile confidence intervals on RULER and LongBench on both models. 

On RULER for Llama-3.1-8B-Instruct (Table~\ref{tab:bootstrap-ruler-llama31-8b}), at the most aggressive compression (25\% budget), CentroidKV achieves $60.26 \pm 0.78$, with a 95\% CI of $\left[58.73, 61.80\right]$, while the next-best methods PyramidKV ($41.59 \pm 0.93$, CI $[39.74, 43.42]$) and SnapKV ($41.78 \pm 0.94$, CI $[39.92, 43.62]$). The confidence intervals are entirely non-overlapping, confirming that the advantage is statistically significant. This separation is consistent at different compression ratios and models shown in Table~\ref{tab:bootstrap-ruler-mistral7b}.

On LongBench with Llama-3.1-8B-Instruct at 25\% budget (Table~\ref{tab:bootstrap-longbench-llama31-8b}), CentroidKV performs comparably to PyramidKV and SnapKV, with overlapping CIs indicating no statistically significant difference among the top three. On Mistral (Table~\ref{tab:bootstrap-longbench-mistral7b}), a similar pattern emerges at moderate cache budget. While for aggressive compression, as analyzed in the main body of paper, the performance degradation of CentroidKV comes from several retrieval-intensive tasks.





\end{document}